\documentclass{article}

    \usepackage[preprint]{neurips_2025}

\usepackage{bbm}
\usepackage{makecell}
\usepackage{hyperref}
\usepackage{float}
\usepackage{booktabs}
\usepackage{adjustbox}
\usepackage{multirow}
\usepackage{wrapfig}
\usepackage{natbib}
\usepackage{microtype}
\usepackage{xcolor}         
\usepackage{amsmath}
\usepackage{amssymb}
\usepackage{multirow}
\usepackage{multicol}
\usepackage{wrapfig}
\usepackage{tabularx}
\usepackage{booktabs}       
\usepackage{amsfonts}       
\usepackage{graphicx}
\usepackage{nicefrac}  
\usepackage{pifont}
\usepackage{colortbl}
\usepackage{longtable}
\usepackage{ulem}
\usepackage{listings}
\usepackage{xspace}
\usepackage{fancyvrb}
\usepackage{xcolor}
\usepackage[most]{tcolorbox}
\usepackage[utf8]{inputenc} 
\usepackage[T1]{fontenc}    
\usepackage{hyperref}       
\usepackage{url}            
\usepackage{soul}
\usepackage{booktabs}       
\usepackage{amsfonts}       
\usepackage{nicefrac}       
\usepackage{microtype}      
\usepackage{xcolor}         

\usepackage{amsmath} 
\usepackage{enumitem} 
\usepackage[utf8]{inputenc} 
\usepackage[T1]{fontenc}    
\usepackage{hyperref}       
\usepackage{url}            
\usepackage{booktabs}       
\usepackage{amsfonts}       
\usepackage{nicefrac}       
\usepackage{microtype}      
\usepackage{xcolor}         
\usepackage{amsmath}
\usepackage{amssymb}
\usepackage[most]{tcolorbox}
\definecolor{lavender}{rgb}{0.882, 0.839, 0.906}
\definecolor{lightyellow}{rgb}{0.98, 0.973, 0.812}
\definecolor{darkyellow}{rgb}{0.98, 0.945, 0.757}
\definecolor{lightpink}{rgb}{1.0, 0.923, 0.913}
\definecolor{darkpink}{rgb}{0.925, 0.769, 0.761}
\usepackage{xcolor}
\definecolor{grey}{rgb}{0.5, 0.5, 0.5}  

\title{Extended Inductive Reasoning for Personalized Preference Inference from Behavioral Signals}

\author{%
  Jia-Nan~Li$^{1~2}$\thanks{Equal contribution.} 
  \quad Jian~Guan$^{2}$\footnotemark[1]
  \quad Wei~Wu$^{2}$\thanks{Corresponding authors: Wei~Wu (wuwei19850318@gmail.com) and Rui~Yan (ruiyan@ruc.edu.cn).}
  \quad Rui~Yan$^{1}$\footnotemark[2]
  \\
  $^{1}$ Gaoling School of Artificial Intelligence, Renmin University of China
  \quad $^{2}$ Ant Group\\
  \texttt{\{lijianan, ruiyan\}@ruc.edu.cn} \\
  \texttt{\{jianguanthu, wuwei19850318\}@gmail.com}\\
}

\begin{document}

\maketitle

\begin{abstract}
Large language models (LLMs) have demonstrated significant success in complex reasoning tasks such as math and coding. In contrast to these tasks where deductive reasoning predominates, inductive reasoning\textemdash the ability to derive general rules from incomplete evidence, remains underexplored. This paper investigates extended inductive reasoning in LLMs through the lens of personalized preference inference, a critical challenge in LLM alignment where current approaches struggle to capture diverse user preferences. The task demands strong inductive reasoning capabilities as user preferences are typically embedded implicitly across various interaction forms, requiring models to synthesize consistent preference patterns from scattered signals. We propose \textsc{AlignXplore}, a model that leverages extended reasoning chains to enable systematic preference inference from behavioral signals in users' interaction histories. Such explicit preference articulation enables efficient streaming inference: when new behavioral signals emerge, the model can directly build upon previously inferred preference descriptions rather than reprocessing historical signals from scratch, while also supporting iterative refinement to the inferred preferences. We develop \textsc{AlignXplore} by combining cold-start training based on synthetic data with subsequent online reinforcement learning. Through extensive experiments, we demonstrate that \textsc{AlignXplore} achieves substantial improvements over the backbone model by an average of 15.49\% on in-domain and out-of-domain benchmarks, while maintaining strong generalization ability across different input formats and downstream models. Further analyses establish best practices for preference inference learning through  systematic comparison of reward modeling strategies, while revealing the emergence of human-like inductive reasoning patterns during training. 







\end{abstract}

\section{Introduction}
Recent advances in large language models (LLMs) have demonstrated remarkable success in complex reasoning tasks through extended reasoning chains~\cite{openai2024o1preview,deepseekai2025deepseekr1incentivizingreasoningcapability}, particularly in domains such as code generation~\cite{chen2021evaluatinglargelanguagemodels} and mathematical problem-solving~\cite{cot,lightman2023let} where deductive reasoning predominates~\cite{morsanyi2018link,chollet2019measure}. However, inductive reasoning, i.e., the ability to derive rules from specific observations and make predictions about novel cases~\cite{hayes2010inductive}, presents unique challenges in probabilistic generalizations from incomplete evidence. As a core cognitive ability~\cite{heit2000properties}, inductive reasoning has long been a key component in human intelligence tests~\cite{ferrara1986children} and scientific research~\cite{kinshuk2006cognitive}. Nevertheless, the extension of LLMs' reasoning abilities to complex inductive tasks remains largely unexplored.

\begin{figure}
    \centering
    \includegraphics[width=\linewidth]{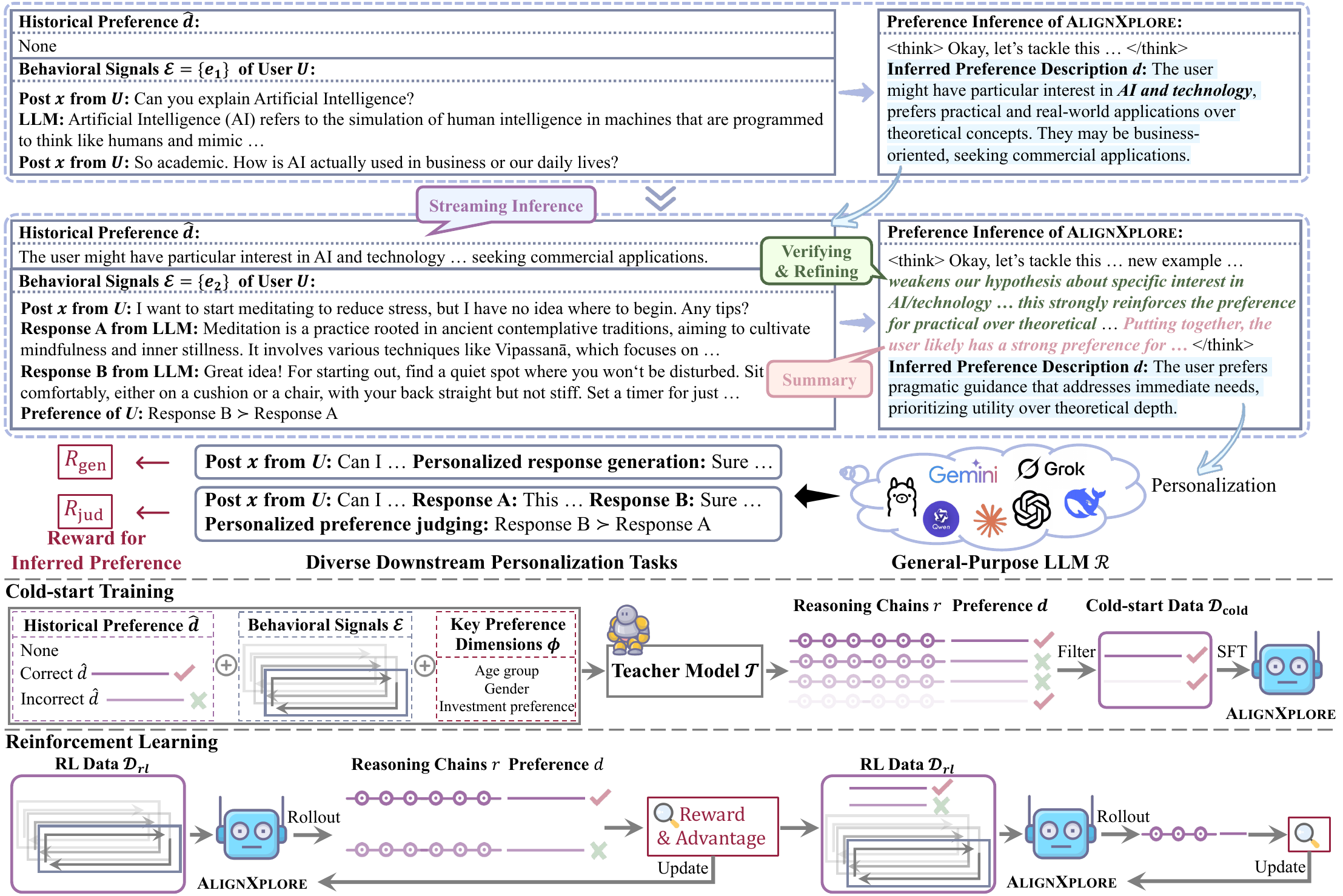}
    \caption{\textbf{Top:} Preference inference task overview. Our model performs human-like inductive reasoning for preference inference by progressively refining their preference hypotheses through iterative testing and validation.  These inferred preferences can then guide diverse downstream personalization tasks. \textbf{Bottom:} Two-stage training process of \textsc{AlignXplore}, which combines cold-start training using synthetic data from teacher models with reinforcement learning optimization to enhance the model's reasoning capabilities.}
    \label{fig:framework}
\end{figure}

In this work, we investigate extended inductive reasoning through the lens of personalized preference inference~\cite{li20251000000usersuserscaling,zhao2025do}, a challenging task that demands strong inductive capabilities to synthesize explicit preference patterns from implicit signals for aligning LLMs with individual preferences. The importance of this investigation is twofold: First, preference inference addresses a critical challenge in LLM alignment, where current approaches primarily focus on universal values such as helpfulness, honesty, and harmlessness~\cite{askell2021general,ouyang2022training,bai2022training,achiam2023gpt,qwen2.5} while struggling to capture the diversity of individual user preferences~\cite{kirk2023the}. This limitation has led to reduced user satisfaction and potential systematic biases~\cite{siththaranjan2024distributional,guan2025surveypersonalizedalignment}, particularly when serving diverse user populations~\cite{tong2023exclusive}. 
Second, preference inference exemplifies the complexities of inductive reasoning. In reality, users rarely explicitly express their preferences, i.e., positive or negative stances towards specific attributes such as cultural sensitivity, during interactions with LLMs~\cite{lee2024aligning}. Instead, these preferences are implicitly embedded in various forms of user-generated content~(e.g., user posts~\cite{wu-etal-2025-aligning}), behavioral signals~(e.g., comparative judgments~\cite{ouyang2022training}), and demographic attributes~(e.g., age, gender~\cite{zhang2018personalizing}). Preference inference requires models to identify consistent preference patterns across such multiple diverse interactions and generalize them to novel contexts, as exemplified in Figure~\ref{fig:framework}. 

Despite the critical importance of preference inference, most existing personalization approaches bypass this crucial step, opting instead for direct mappings that incorporate implicit signals as prompts~\cite{xu2022beyond,lee2024aligning}, trainable parameters~\cite{kang2023llms,tan2024democratizing}, or encoded hidden representations~\cite{poddar2024personalizing,ning2024user}. The absence of explicit preference inference renders the preference modeling process opaque and uncontrollable. Worse yet, without explicit preference representations, these methods cannot incrementally refine user preferences when new behavioral signals become available, forcing models to process growing interaction histories from scratch and limiting efficiency. To address these challenges, we propose \textsc{AlignXplore}, a model that leverages extended reasoning chains to enable systematic inductive reasoning from behavioral signals. Through explicit preference articulation, our model naturally enables a streaming inference mechanism that incrementally incorporates previously inferred preferences for more efficient personalization. To this end, we develop a two-stage framework that combines synthetic data training with reinforcement learning optimization. First, we address the cold-start challenge by leveraging advanced LLMs to generate high-quality training data that demonstrates the process of preference inference through extended reasoning. We then enhance the model's reasoning capabilities through reinforcement learning, where the reward signal is designed to encourage accurate preference inference.


Through extensive experiments on both in-domain and out-of-domain benchmarks, we demonstrate that \textsc{AlignXplore} achieves substantial improvements in personalized alignment, outperforming the backbone model by 15.49\% and showing competitive performance against significantly larger models, including GPT-4~\cite{achiam2023gpt} and DeepSeek-R1-671B~\cite{deepseekai2025deepseekr1incentivizingreasoningcapability}. Notably, \textsc{AlignXplore}'s streaming inference mechanism enables efficient processing of growing behavioral signals by avoiding repeated computation, while simultaneously allowing for the gradual refinement of inferred preferences, thereby achieving better performance. \textsc{AlignXplore} also demonstrates strong generalization ability across different input formats and downstream models, and maintains robust performance under preference reversal. This is attributed to the extended reasoning process that helps the model develop more systematic and transferable inductive reasoning patterns, rather than learning superficial correlations. Further analysis reveals two key findings: (1) comparing different reward modeling approaches reveals that directly optimizing for preference judging leads to more stable training than optimizing response generation, establishing best practices for training preference inference models, and (2) our two-stage training approach demonstrates progressive enhancement of inductive reasoning capabilities, where cold-start training helps establish basic preference characterization abilities, while RL further refines these into actionable hypotheses through iterative testing and refinement, mirroring human approaches to inductive reasoning~\cite{heit2000properties}.

The main contributions of this work are as follows:

I. We present the first systematic investigation of extended inductive reasoning in LLMs through the lens of personalized preference inference, demonstrating how structured reasoning processes enable LLMs to derive generalizable preference patterns from implicit behavioral signals.

II. We develop \textsc{AlignXplore}, a preference inference model that naturally supports efficient processing of growing behavioral signals through streaming inference. Our model is trained with a novel two-stage framework combining synthetic data training with reinforcement learning, and we open-source our implementation to facilitate future research in personalized alignment.\footnote{Code is available at \url{https://github.com/AntResearchNLP/AlignXplore}.}

III. We conduct comprehensive evaluations across diverse benchmarks, demonstrating substantial improvements over existing approaches while maintaining strong efficiency, generalization ability and robustness. Our analyses provide valuable insights into reward modeling strategies and the progressive development of inductive reasoning capabilities.


\section{Related works}

\paragraph{Inductive reasoning}

Unlike deductive reasoning, where conclusions follow deterministically from premises, inductive reasoning involves making probabilistic generalizations from incomplete evidence~\cite{Lake_Ullman_Tenenbaum_Gershman_2017,hayes2010inductive}, which is crucial for various cognitive activities from categorization to scientific discovery~\cite{holland1986induction}. This capability has gained renewed attention through the Abstract Reasoning Corpus (ARC)~\cite{chollet2019measure,moskvichev2023the} in evaluating LLMs like OpenAI o3~\cite{pfister2025understandingbenchmarkingartificialintelligence}. While existing research~\cite{wang2024hypothesis} primarily focuses on few-shot generalization~\cite{radford2018improving,brown2020language}, preference inference presents three distinct challenges: reasoning over unstructured language instead of formal languages~\cite{qiu2024phenomenal,yanmir}, handling heterogeneous forms of preference signals that may significantly differ from test-time user tasks, and necessitating reasoning about negative examples that reveal undesired preferences~\cite{laskin2023incontext}. Our framework provides a principled solution to these challenges while maintaining interpretability.

\paragraph{Extended reasoning in LLMs} Traditional Chain-of-Thought approaches~\cite{cot} are limited by shallow, linear reasoning steps. Recent advances in extended reasoning~\cite{openai2024o1preview,chen2025towards} have significantly improved LLMs' performance through three key mechanisms: (1) In-depth logical chains that maintain extended reasoning through various formats, including natural language~\cite{wang2planning}, formal language~\cite{wen2024codeplan}, and latent space reasoning~\cite{hao2024traininglargelanguagemodels}; (2) Systematic exploration of solution spaces, implemented via internal mechanisms trained by reinforcement learning~\cite{deepseekai2025deepseekr1incentivizingreasoningcapability} or external frameworks like Monte Carlo tree search~\cite{zhang2024rest} and beam search~\cite{tot,snell2025scaling}; and (3) Iterative self-reflection that enables models to verify and correct reasoning paths through supervised fine-tuning~\cite{qwen2.5,gandhi2025cognitive} or reinforcement learning with verifiable rewards~\cite{deepseekai2025deepseekr1incentivizingreasoningcapability,yu2025dapo}. The integration of these mechanisms has led to significant improvements in complex reasoning tasks such as math~\cite{hu2025open}, coding~\cite{jainlivecodebench}, scientific question-answering~\cite{rein2024gpqa}, reward modeling~\cite{chen2025rmr1rewardmodelingreasoning}, and multimodal reasoning~\cite{shen2025vlm}. We extend this paradigm to preference inference, a domain that poses unique challenges due to its requirement for strong inductive reasoning capabilities. 

\paragraph{Personalized alignment} Recent studies highlight limitations of one-size-fits-all alignment~\cite{askell2021general,kirk2023the,siththaranjan2024distributional}, motivating personalized alignment,  i.e., adapting LLM behaviors to individual preferences~\cite{kalai1975other,oldenburg2024learning}. Key challenges include: (1) Preference inference from implicit signals~\cite{wu-etal-2025-aligning,ouyang2022training,zhang2018personalizing}, which requires sophisticated reasoning to synthesize scattered signals~\cite{guan2025surveypersonalizedalignment}. Current works primarily focus on retrieving preference-relevant contexts~\cite{zhao2025do,pan2025secom,10.1145/3604915.3608885} while overlooking explicit preference inference, leading to limited alignment accuracy~\cite{li20251000000usersuserscaling}. (2) Preference modeling through prompts~\cite{xu2022beyond,lee2024aligning}, model parameters~\cite{kang2023llms,tan2024democratizing}, or latent representations~\cite{poddar2024personalizing,ning2024user}. We focus on prompt-based methods for their interpretability and model-agnostic nature. (3) Feedback-driven alignment that updates LLMs during training~\cite{jang2023personalized,guan2024amor,kuang2024federatedscope} or guides generation at inference~\cite{shi2024decoding,chen2025pal,rame2023rewarded}. In contrast to existing approaches, we present the first study incorporating both extended reasoning for accurate preference inference and efficient mechanisms for handling evolving user interactions~\cite{chandrashekaran1996modeling}.

\section{Methodology}
In this section, we first formulate the preference inference task and evaluation methods in \S\ref{formulation}, then detail our two-phase training strategy to develop the preference inference model: an initial cold-start phase to develop basic reasoning capabilities~\ref{cold_train}, followed by a reinforcement learning phase that directly optimizes for the reward~\ref{rl_train}. Figure~\ref{fig:framework} illustrates the training recipe.

\subsection{Task formulation}\label{formulation}
We first formulate the preference inference task as follows:
Given a collection of behavioral signals $\mathcal{E} = \{e_1, e_2, ..., e_T\}$ with multiple interaction examples of user $U$,\footnote{For simplicity, our main experiments use comparative judgments (a user post with preferred/less-preferred responses) as preference signals, though our method is agnostic to both the source platforms and signal formats, readily accommodating various forms of implicit signals such as user posts, reviews, or interaction histories.} the model $\mathcal{M}$ should generate an explicit preference description $d$ in natural language with an extended reasoning chain $r$:
\begin{align}
    r, d=\mathcal{M}(\mathcal{E}),
\end{align}
where $d$ typically manifests as positive or negative attitudes of $U$ towards specific dimensions (e.g., cultural sensitivity, formality, etc.). The inferred preference description $d$ should be model-agnostic, enabling it to condition any general-purpose LLM $\mathcal{R}$ for personalization realization~\cite{lee2024aligning,li20251000000usersuserscaling}.

\paragraph{Streaming inference mechanism} In real-world scenarios, user behavioral signals are continuously updated over time, often accumulating to large volumes of data. To address this computational efficiency challenge, we propose a novel streaming inference mechanism that incrementally leverages previously inferred preferences. Our key insight is that as new behavioral signals $\mathcal{E}$ become available during user interactions, we can perform efficient inference by conditioning on the previously inferred preference description $\hat{d}$ rather than re-encoding the corresponding out-of-date behavioral signals:
\begin{align}
    r, d=\mathcal{M}(\mathcal{E}, \hat{d}),
\end{align}
where $\hat{d}$ is the historical preference that can be viewed as a condensed representation of all behavioral signals before $\mathcal{E}$. This streaming mechanism is uniquely enabled by explicit preference modeling, whereas previous methods that rely on prompting~\cite{xu2022beyond,li20251000000usersuserscaling} or parameter updates~\cite{kang2023llms,poddar2024personalizing} need to process the entire historical interactions whenever personalization is needed for downstream tasks.

\paragraph{Evaluation framework} The quality of generated preference $d$ can be assessed by how well it guides $\mathcal{R}$ to align with user preferences. Ideally, this could be measured through an online reward:
\begin{align}
    R_{\rm online}=\mathbb{E}_{o\in \mathcal{R}(\cdot|x,d)}\text{Align}(o, U),
\end{align}
where $o$ represents $\mathcal{R}$'s output on a new post $x$ of user $U$ and $\text{Align}(\cdot)$ measures its alignment with the user. However, this approach requires costly online sampling and user feedback. To enable efficient and scalable evaluation while avoiding such overhead, we leverage offline user-specific comparative judgment data. Specifically, given a post $x$ from user $U$ and two responses $y_w$ and $y_l$ where $y_w$ is preferred over $y_l$ by $U$, we define:
\begin{align}
\label{reward}
    R_{\rm offline}&=\mathbbm{1}\big(f_\mathcal{R}(y_w|x, \cdot)> f_\mathcal{R}(y_l|x, \cdot)\big)R_{\rm format},\\
    R_{\rm format}&=\mathbbm{1}\big(r, d\text{~satisfy the generation format}\big),\label{reward_format}
\end{align}
where $f_\mathcal{R}(y_{w/l}|x, \cdot)$ measures the model's preference scores for the two responses, and $R_{\rm format}$ ensures the structural validity of both $r$ and $d$ (see Appendix \ref{appendix:prompt} for format specifications).

\paragraph{Reward instantiation} The above offline evaluation framework can be instantiated by modeling preference scores $f_\mathcal{R}(y_{w/l}|x, \cdot)$ in different ways. For example, when the downstream model $\mathcal{R}$ is repurposed as a response generation model (denoted as $\mathcal{R}_{\rm gen}$)~\cite{rafailov2024direct}, it measures preference of a response $y_{w/l}$ through the change in response log-probability when conditioned on $d$ compared to the unconditional case. The offline reward, denoted as $R_{\rm gen}$, then compares the log-probability changes between $y_w$ and $y_l$, where a larger positive margin indicates better preference alignment:
\begin{align}
    R_{\rm gen} &= \mathbbm{1}\big(\log\frac{\mathcal{R}_{\rm gen}(y_w|x,d)}{\mathcal{R}_{\rm gen}(y_w|x)}>\log\frac{\mathcal{R}_{\rm gen}(y_l|x,d)}{\mathcal{R}_{\rm gen}(y_l|x)}\big)R_{\rm format}.\label{reward_gen}
\end{align}
When $\mathcal{R}$ serves as a preference judging model (denoted as $\mathcal{R}_{\rm jud}$)~\cite{zheng2023judging}, it directly models the preference score using the probability of a response $y_{w/l}$ being preferred under the inferred preference description $d$. The corresponding offline reward, denoted as $R_{\rm jud}$, is computed based on the probability difference between $y_w$ and $y_l$. Specifically:
\begin{align}
    R_{\rm jud} &= \mathbbm{1}\big(\mathcal{R}_{\rm jud}(y_w|x,d,y_w,y_l) > \mathcal{R}_{\rm jud}(y_l|x,d,y_w,y_l)\big)R_{\rm format}.\label{reward_judge}
\end{align}
Our evaluation framework can be further instantiated with other types of $\mathcal{R}$, such as using raw response log-probabilities directly as preference scores~\cite{meng2024simpo}. We leave the exploration of these alternative reward formulations as future work. In our main experiments, we primarily use $R_{\rm jud}$ for both training and evaluation, while analyzing $R_{\rm gen}$ in subsequent ablation studies.

\subsection{Cold-start training}\label{cold_train}
The primary challenge in training preference inference models lies in the inherent difficulty for small models to perform complex preference inference following instructions alone without proper initialization. To address this, we develop a synthetic data generation pipeline leveraging advanced LLMs to create high-quality training examples with detailed reasoning processes.

Specifically, we employ a two-stage data synthesis process. For each example in the original implicit preference signals $e_i\in\mathcal{E}$, we first identify key preference dimensions $\phi$ expressed in natural language that potentially reveal user preferences, which serve as analytical guidance for subsequent preference inference. We then prompt an advanced teacher model $\mathcal{T}$ with both these identified dimensions $\phi$ and the original implicit signals to generate $G$ reasoning chains and preference descriptions (see Appendix \ref{appendix:prompt} for prompt templates):
    $\{r_i,d_i\}_{i=1}^G\sim \mathcal{T}(r,d|\mathcal{E},\phi).$
To support streaming inference, we construct training examples that simulate the incremental preference learning process. Specifically, for each user, we first randomly select a previously generated preference description $\hat{d}$ as historical preference, then sample a new set of behavioral signals from the same user. These pairs of historical preferences and new signals are used to condition the teacher model $\mathcal{T}$ for generating new reasoning chains and updated preference descriptions. After mixing these streaming examples with the original ones, we filter all generations through outcome-based verification, selecting only the samples that achieve optimal reward scores. The filtered dataset $\mathcal{D}_{\text{cold}}$ is constructed as:
\begin{align}
    \mathcal{D}_{\text{cold}} = \{(\mathcal{E},\hat{d},r_i,d_i)| R(r_i,d_i)=1, i\in[1,G]\},
\end{align}
where $R(\cdot)$ denotes either $R_{\text{gen}}$ or $R_{\text{judge}}$ as defined in Equations \ref{reward_gen} and \ref{reward_judge}, respectively, and $\hat{d}_i$ may be empty for examples without historical preferences. The training objective of the preference inference model $\mathcal{M}$ is to maximize the likelihood of generating both correct reasoning chains and accurate preference descriptions:
\begin{align}
    \mathcal{L}_{\text{cold}} = \mathbb{E}_{(\mathcal{E},\hat{d}, r,d)\sim \mathcal{D}_{\rm cold}} -\frac{1}{|r|+|d|}\sum_{t=1}^T\log p(r,d|\mathcal{E},\hat{d}),
\end{align}
where $p(\cdot|\mathcal{E}, \hat{d})$ denotes the conditional probability distribution modeled by $\mathcal{M}$.

\subsection{Reinforcement learning}\label{rl_train}
While cold-start training establishes basic reasoning capabilities, RL further enhances the model's ability to generate high-quality preference descriptions through extended reasoning. We adopt the Group Relative Policy Optimization (GRPO) algorithm~\cite{deepseekai2025deepseekr1incentivizingreasoningcapability}, which has demonstrated effectiveness in optimizing long-horizon reasoning processes. Specifically, for each training instance, we sample multiple reasoning paths and optimize them collectively using the reward signal defined in Eq.~\ref{reward}. Following~\cite{hu2025open}, we remove the KL penalty term from the original GRPO formulation for more effective optimization:
\begin{align}
    \mathcal{L}_\text{RL} &= \mathbb{E}_{\substack{(\mathcal{E},\hat{d})\sim\mathcal{D}_\text{rl}\\ \{(r_{i},d_{i})\}_{i=1}^G\sim p_{\text{old}}(\cdot|\mathcal{E},\hat{d})}} \big(-\frac1G\sum_{i=1}^G\frac{1}{|r_i|+|d_i|}\rho_i\big),\\
    \rho_{i}&=\sum_{t}\min\big(\frac{p(\{r_{i},d_{i}\}_t|\mathcal{E},\hat{d})}{p_{\text{old}}(\{r_{i},d_{i}\}_t|\mathcal{E},\hat{d})} A_{i}, \text{clip}(\frac{p(\{r_{i},d_{i}\}_t|\mathcal{E},\hat{d})}{p_{\text{old}}(\{r_{i},d_{i}\}_t|\mathcal{E},\hat{d})}, 1-\epsilon, 1+\epsilon)A_{i}\big),\\
    A_{i}&=\frac{R_i-\text{mean}(\{R_j\}_{j=1}^G)}{\text{std}(\{R_j\}_{j=1}^G)},
\end{align}
where $p_{\text{old}}$ is the old policy model, $G$ is the number of sampled outputs, $\{r_{i},d_{i}\}_t$ is the $t$-th token in the generated sequence, and $R_i$ is the reward of $i$-th output, computed using Eq.~\ref{reward_gen} or~\ref{reward_judge}. The advantage term $A_{i}$ normalizes rewards across different paths to reduce variance in training. The historical preference $\hat{d}$ is either empty or dynamically generated by $p_{\text{old}}$ using other behavioral signals from the same user as $\mathcal{E}$.

\section{Experiments}

\subsection{Experimental setup}
\label{Experimental_setup}
\paragraph{Implementation details}
We maintain a streaming setting for training. Specifically, during both cold-start data generation and RL training rollouts, we first generate preference descriptions with empty historical preference $\hat{d}$ using 4 behavioral examples ($T=4$ in $\mathcal{E}$) for each training instance. Then, we randomly sample another 4 examples from the same user as new behavioral signals $\mathcal{E}$, and use the previously inferred preference as $\hat{d}$ to generate second-round preferences. The training data combines instances from both rounds. While our experiments demonstrate this two-round setup, the streaming mechanism naturally extends to more rounds by iteratively using inferred preferences as historical preferences. We adopt DeepSeek-R1-Distill-Qwen-7B~\cite{deepseekai2025deepseekr1incentivizingreasoningcapability} as our backbone model and conduct training on \textsc{AlignX}~\cite{li20251000000usersuserscaling}, a comprehensive personalized alignment dataset spanning 90 preference dimensions with balanced positive and negative examples. We create two separate training sets from \textsc{AlignX}: 7000 instances for cold-start training, along with 7,000 instances for RL, using $R_\text{jud}$~(Eq.~\ref{reward_judge}) as the reward function and QwQ-32B~\cite{qwq32b} as the teacher model. For RL training, we use a prompt batch size of 128 with 4 rollouts per prompt. During inference, we combine nucleus sampling ($p=0.95$)~\cite{holtzmancurious} with top-$k$ sampling ($k=10$)~\cite{fan2018hierarchical} and set the temperature to 0.9~\cite{goodfellow2014generative}. For in-depth analysis on the impact of different configurations, we further train two models in a base setting where $\hat{d}$ is always empty and $\mathcal{E}$ contains either 4 or 8 examples. Appendix~\ref{implementation_details} shows more implementation details.

\paragraph{Benchmarks}
We evaluate on two benchmarks: (1) \textsc{AlignX}$_{\rm test}$~\citep{li20251000000usersuserscaling}, the official test set of \textsc{AlignX}; and (2) \textsc{P-Soups}~\citep{jang2023personalized}, which focuses on three preference dimensions: ``expertise,'' ``informativeness,'' and ``style.'' Table~\ref{tab:benchmark} summarizes the statistics. Following our training settings, we consider two evaluation settings: a base setting where models perform inference using 4 or 8 preference pairs with empty $\hat{d}$, and a streaming setting where the model first uses the initial 4 pairs to infer a historical preference description $\hat{d}$, then combine $\hat{d}$ with 4 new pairs to generate the final preference description. All preference pairs are randomly sampled from the same user's behavioral signals. We ensure that each model is evaluated under its corresponding training setting.





\begin{table}[!t]
    \centering
    \caption{Summary of evaluation benchmarks. For preference directions, $\uparrow$ and $\downarrow$ represent preferred and non-preferred examples, respectively, with their quantities shown in parentheses. The ``In-domain'' column (\ding{51}/\ding{55}) indicates whether the benchmark's preference dimensions are seen during training.}
    \begin{adjustbox}{max width=\linewidth}
    \begin{tabular}{@{}llc@{}}
    \toprule
    \textbf{Benchmark} & \multicolumn{1}{c}{\textbf{Dimensions and \#Examples}} & \textbf{In-domain} \\
    \midrule
    \textbf{\textsc{AlignX}$_{\rm test}$}&90 preference dimensions (3,000 examples in total, $\sim$1:1 ratio for $\uparrow$/$\downarrow$ preferences)&\ding{51}\\
    \textbf{P-\textsc{Soups}}   & ``Expertise'' ($\uparrow$: 300, $\downarrow$: 300); ``Informativeness'' ($\uparrow$: 300, $\downarrow$: 300); ``Style'' ($\uparrow$: 300, $\downarrow$: 300) &\ding{55}\\
    \bottomrule
    \end{tabular}
    \end{adjustbox}
    \label{tab:benchmark}
\end{table}

\begin{table}[t]
\centering
\small
\caption{Offline preference inference evaluation results (ACC$_\text{jud}$, \%) using Qwen2.5-7B-Instruct as the preference judging model. \textbf{Extended Reasoning}: whether the model generates preference descriptions with extended reasoning. 
All non-pink-shaded rows represent results with only 4 preference pairs available in $\mathcal{E}$, while  \sethlcolor{pink!20}\hl{pink-shaded} rows show results where 4 additional preference pairs are given. $T$ refers to the number of examples in $\mathcal{E}$ in both training and inference. $\hat{d}$ indicates whether historical preferences are empty or not. \sethlcolor{gray!20}\hl{gray-shaded} rows represent golden preference or large-sized models, where \textit{\textcolor{gray}{italic}} numbers indicate performance below the best result (in \textbf{bold}) of non-gray-shaded methods. \underline{Underlined} numbers indicate the second-best results. * indicates that the best result is significantly better than the others ($p < 0.05$ with pairwise $t$-test).}
\begin{adjustbox}{max width=\linewidth}
\begin{tabular}{lcccccccc}
\toprule
\multirow{2}{*}{\textbf{Method}} & \multirow{1}{*}{\textbf{Extended}} &\multirow{2}{*}{\textbf{Setting}} &\multirow{2}{*}{$T$} & \multirow{2}{*}{$\hat{d}$} & \multirow{2}{*}{\textsc{\textbf{AlignX}}$_\text{test}$} & \multicolumn{3}{c}{\textsc{\textbf{P-Soups}}}\\
\cmidrule{7-9}
&\textbf{Reasoning}&&&&&\textbf{Informativeness} & \textbf{Style} & \textbf{Expertise} \\
\midrule
\midrule
\multicolumn{9}{c}{\textbf{\textit{Directly given preference descriptions}}}\\
\midrule
\textbf{Null}&N/A&N/A&N/A&N/A& ~~51.37$^*$ & ~~45.85$^*$ & ~~17.00$^*$ & ~~36.00$^*$\\
${\mathcal{E}}$&N/A&N/A&4&N/A& ~~50.33$^*$ & ~~41.03$^*$ & ~~37.33$^*$ & ~~36.00$^*$\\
\rowcolor{gray!20}\textbf{Golden Preference}&N/A&N/A &N/A&N/A& \textcolor{gray}{\textit{64.63}} & 68.94 & 84.50 & 90.17\\
\midrule
\midrule
\multicolumn{9}{c}{\textbf{\textit{Previous specialized methods for inductive reasoning and personalization}}}\\
\midrule
\textbf{LMInductReason}~\cite{qiu2024phenomenal}&N/A&Base &4&\ding{55}& ~~51.80$^*$ & ~~44.35$^*$ & ~~27.50$^*$ & ~~38.17$^*$\\
\textbf{VPL}~\cite{poddar2024personalizing}&N/A &Base&4&\ding{55}& ~~51.20$^*$ & ~~43.69$^*$ & ~~47.17$^*$ & ~~52.67$^*$\\
\textbf{PBA}~\cite{li20251000000usersuserscaling}&N/A&Base &4&\ding{55}& ~~{62.77}$^*$ & ~~53.65$^*$ & ~~31.33$^*$ & ~~50.50$^*$\\
\midrule
\midrule
\multicolumn{9}{c}{\textbf{\textit{Preference descriptions generated by state-of-the-art LLMs}}}\\
\midrule
\textbf{Qwen2.5-7B-Instruct}~\cite{qwen2.5}&\ding{55} &Base&4&\ding{55}& ~~56.33$^*$ & ~~{53.82}$^*$ & ~~59.00$^*$ & ~~{65.17}$^*$\\
\textbf{DS-R1-Distill-Qwen-7B}~\cite{deepseekai2025deepseekr1incentivizingreasoningcapability}&\ding{51}&Base &4&\ding{55}& ~~57.63$^*$ & ~~51.16$^*$ & ~~45.83$^*$ & ~~56.67$^*$\\
\rowcolor{pink!20}\textbf{DS-R1-Distill-Qwen-7B}~\cite{deepseekai2025deepseekr1incentivizingreasoningcapability}&\ding{51}&Base &8&\ding{55}& ~~56.13$^*$ & ~~49.50$^*$ & ~~56.17$^*$ & ~~57.33$^*$\\
\rowcolor{pink!20}\textbf{DS-R1-Distill-Qwen-7B}~\cite{deepseekai2025deepseekr1incentivizingreasoningcapability}&\ding{51}&Streaming &4&\ding{51}& ~~56.40$^*$ & ~~50.00$^*$ & ~~49.50$^*$ & ~~60.17$^*$\\
\midrule
\rowcolor{gray!20}\textbf{Qwen3-32B$_\text{non-thinking}$}~\cite{qwen2025qwen3}&\ding{55} &Base&4&\ding{55}& \textcolor{gray}{\textit{57.60}} & \textcolor{gray}{\textit{54.98}} & \textcolor{gray}{\textit{61.50}} & \textcolor{gray}{\textit{66.67}}\\
\rowcolor{gray!20}\textbf{GPT-4}~\cite{achiam2023gpt}&\ding{55} &Base&4&\ding{55}& \textcolor{gray}{\textit{66.10}} & \textcolor{gray}{\textit{53.82}} & \textcolor{gray}{\textit{73.33}} & 71.83\\
\rowcolor{gray!20}\textbf{QwQ-32B}~\cite{qwq32b}&\ding{51}&Base &4&\ding{55}& \textcolor{gray}{\textit{65.70}} & \textcolor{gray}{\textit{58.14}} & \textcolor{gray}{\textit{72.17}} & 71.50\\
\rowcolor{gray!20}\textbf{Qwen3-32B$_\text{thinking}$}~\cite{qwen2025qwen3}&\ding{51}&Base&4&\ding{55}& \textcolor{gray}{\textit{65.03}} & \textcolor{gray}{\textit{57.14}} & \textcolor{gray}{\textit{71.67}} & 73.83\\
\rowcolor{gray!20}\textbf{DeepSeek-R1-671B}~\cite{deepseekai2025deepseekr1incentivizingreasoningcapability}&\ding{51} &Base&4&\ding{55}& \textcolor{gray}{\textit{70.47}} & \textcolor{gray}{\textit{55.48}} & \textcolor{gray}{\textit{79.66}} & 76.17\\
\rowcolor{pink!20}\textbf{DeepSeek-R1-671B}~\cite{deepseekai2025deepseekr1incentivizingreasoningcapability}&\ding{51} &Base&8&\ding{55}&\textcolor{gray}{\textit{70.23}} & \textcolor{gray}{\textit{56.98}} & 84.17 & 79.17\\
\rowcolor{pink!20}\textbf{DeepSeek-R1-671B}~\cite{deepseekai2025deepseekr1incentivizingreasoningcapability}&\ding{51} &Streaming&4&\ding{51}&\textcolor{gray}{\textit{67.70}} & \textcolor{gray}{\textit{56.64}} & \textcolor{gray}{\textit{69.50}} & \textcolor{gray}{\textit{69.17}}\\
\midrule
\midrule
\multicolumn{9}{c}{\textbf{\textit{Preference descriptions generated by our preference inference model}}}\\
\midrule
\textbf{\textsc{AlignXplore}-7B}&\ding{51}&Base&4&\ding{55} & \underline{65.33} & {54.32} & {69.67} & {63.83}\\
\rowcolor{pink!20}\textbf{\textsc{AlignXplore}-7B}&\ding{51}&Base&8&\ding{55} & 64.30&\underline{57.14}&\underline{70.33}&\underline{66.50}\\
\rowcolor{pink!20}\textbf{\textsc{AlignXplore}-7B}&\ding{51}&Streaming&4&\ding{51} & \textbf{71.47} & \textbf{61.30} & \textbf{83.00} & \textbf{71.33}\\
\midrule
\textbf{\textsc{AlignXplore}-7B w/o RL}&\ding{51}&Base&4&\ding{55}& {61.80} & 52.82 & 54.00 & 59.83\\
\textbf{\textsc{AlignXplore}-7B w/o Cold-start}&\ding{51}&Base &4&\ding{55}& {62.80} & 56.64 & {64.83} & 59.50\\
\bottomrule
\end{tabular}
\end{adjustbox}
\label{main_results}
\end{table}

\paragraph{Evaluation metrics}
Due to the inherent difficulty of directly evaluating preference inference quality, we employ both offline and online metrics for indirect evaluation: \textbf{(1) Offline evaluation:} We measure \textbf{Acc$_\text{gen}$} and \textbf{Acc$_\text{jud}$} following Eq.~\ref{reward_gen} and \ref{reward_judge}, which assess preference-guided response generation and preference judging accuracy, respectively. We primarily focus on \textbf{Acc$_\text{jud}$} as it aligns with our training objective. \textbf{(2) Online evaluation:} We introduce \textbf{GPT-4 Win Rate},\footnote{We use OpenAI's API ``gpt-4-turbo-2024-04-09'' for our all subsequent experiments.} where GPT-4 conditioned on the ground-truth preferences (provided by the benchmarks) compares responses generated given preference descriptions from different models~\citep{kumar2024compo,jang2023personalized}.

\paragraph{Baselines}
We compare our approach with three groups of baselines: 
    \textbf{(1) Direct preference descriptions:} \textit{Null} (no description), $\mathcal{E}$ (raw behavioral signals), and \textit{Golden Preference} (ground-truth descriptions from benchmarks\footnote{Note that golden preference descriptions, while semantically accurate, may not necessarily lead to optimal downstream personalization performance due to potential gaps in model compatibility.}).   
    \textbf{(2) Specialized methods:} \textit{LMInductReason}~\cite{qiu2024phenomenal} for inductive reasoning, \textit{VPL}~\cite{poddar2024personalizing} for preference modeling, and \textit{PBA}~\cite{li20251000000usersuserscaling} for structured preference prediction.
    \textbf{(3) State-of-the-art LLMs:} Small models (\textit{Qwen2.5-7B-Instruct}~\cite{qwen2.5}, \textit{DS-R1-Distill-Qwen-7B}~\cite{deepseekai2025deepseekr1incentivizingreasoningcapability}) and large models (\textit{QwQ-32B}~\cite{qwq32b}, \textit{Qwen3-32B}~\cite{qwen2025qwen3}, \textit{GPT-4}~\cite{achiam2023gpt}, \textit{DeepSeek-R1-671B}\footnote{All experiments are based on the DeepSeek-R1 version released on 2025/01/20.}~\cite{deepseekai2025deepseekr1incentivizingreasoningcapability}).

We also evaluate ablated versions of our model (w/o RL and w/o Cold-start) to verify the effectiveness of each training stage. See Appendix~\ref{baseline_details} for baseline implementation details.


\subsection{Main results}

\begin{table}[!t]
\caption{Online preference inference evaluation results (GPT-4 win rate, \%, row model against column model) using Qwen2.5-7B-Instruct as the personalized response generation model. We randomly select 400 test cases per benchmark for evaluation. \textbf{M1}: Qwen2.5-7B-Instruct; \textbf{M2}: DS-R1-Distill-Qwen-7B; \textbf{M3}: \textsc{AlignXplore}-7B.}
\label{tab:win_rate}
\centering
\begin{adjustbox}{max width=0.7\columnwidth}
\begin{tabular}{ccccccccc}
    \toprule
    {\cellcolor{lavender} \textbf{\textsc{AlignX}$_{\rm test}$}} & \textbf{M1} & \textbf{M2} & \textbf{M3} &\textbf{~~~~~~~~~}&{\cellcolor{lavender} \textbf{P-\textsc{Soups}}} & \textbf{M1} & \textbf{M2} & \textbf{M3} \\
    \midrule
    \textbf{M1} & - & \cellcolor{darkyellow}43.00 & \cellcolor{lightyellow}37.00 && \textbf{M1} & - & \cellcolor{lightpink}51.33 & \cellcolor{darkyellow}42.33 \\
    \textbf{M2} & \cellcolor{lightpink}57.00 & - & \cellcolor{darkyellow}43.00 && \textbf{M2} & \cellcolor{darkyellow}48.67 & - & \cellcolor{darkyellow}46.67 \\
    \textbf{M3} & \cellcolor{darkpink}63.00 & \cellcolor{lightpink}57.00 & - && \textbf{M3} & \cellcolor{lightpink}57.67 & \cellcolor{lightpink}53.33 & - \\
    \bottomrule
\end{tabular}
\end{adjustbox}
\end{table}

\begin{table}[t]
\centering
\small
\caption{Generalization and robustness evaluation (ACC$_\text{jud}$, \%). \textbf{Generalization:} (1) Input format generalization: inferring preferences from user-generated content (UGC), shown in the ``\textsc{AlignX}$_\text{test}$ w/ UGC'' column; (2) Cross-model transferability: personalizing different preference judging models (columns) using generated preference descriptions (rows) on the original \textsc{AlignX}$_\text{test}$ benchmark. \textbf{Robustness:} Evaluating model performance when preference directions are reversed in both behavioral signals and test pairs. The subscripts in the last two columns indicate performance changes compared to the original results in Table~\ref{main_results}. \textbf{Takeaway:} \textsc{AlignXplore}-7B shows strong generalization abilities in both aspects while maintaining robust performance under preference reversal, suggesting it captures fundamental preference patterns rather than learning fixed biases.}
\begin{adjustbox}{max width=\linewidth}
\begin{tabular}{lcccccm{1pt}cc}
\toprule
\multirow{3}{*}{\textbf{Method}} &\multirow{3}{*}{{\textbf{\makecell{Extended\\Reasoning}}}}& \multicolumn{4}{c}{\textbf{Generalization}} && \multicolumn{2}{c}{\textbf{Robustness}}\\
\cmidrule{3-6}
\cmidrule{8-9}
& & 
\multirow{2}{*}{{\textbf{\makecell{\textsc{\textbf{AlignX}}$_\text{test}$\\\textbf{w/ UGC}}}}}
& \multicolumn{3}{c}{\textbf{Preference Judging Model $\mathcal{R}_\text{jud}$}}&&\multirow{2}{*}{{\textbf{\makecell{\textsc{\textbf{AlignX}}$_\text{test}$\\\textbf{(Reverse)}}}}}&\multirow{2}{*}{{\textbf{\makecell{\textsc{\textbf{P-Soups}}\\\textbf{(Reverse)}}}}} \\
\cmidrule{4-6}
&&&\textbf{Qwen2.5-7B-Instruct} & \textbf{QwQ-32B} & \textbf{DeepSeek-R1-671B} && & \\
\midrule
\midrule
${\mathcal{E}}$&N/A & 52.17 & 50.33 & 49.03 & 50.12&&48.67$_{-1.7}$&36.57$_{-1.6}$\\
\rowcolor{gray!20}\textbf{Golden Preference}&N/A & 69.87 & \textcolor{gray}{\textit{64.63}} & 74.30 & 78.97 && \textit{\textcolor{gray}{61.83$_{-2.8}$}} & \textit{\textcolor{gray}{67.42$_{-13.8}$}}\\
\midrule
\midrule
\textbf{Qwen2.5-7B-Instruct}&\ding{55} & 57.57&56.33 & 56.90 & 58.15&&47.27$_{-9.1}$&\underline{68.33}$_{+9.0}$\\
\textbf{DS-R1-Distill-Qwen-7B}&\ding{51} & \underline{58.30}&\underline{57.63} & \underline{58.70} & \underline{59.61}&&\underline{53.40}$_{-4.2}$&67.83$_{+16.6}$\\
\midrule
\rowcolor{gray!20}\textbf{DeepSeek-R1-671B}&\ding{51} & \textcolor{gray}{\textit{61.97}} &70.47 & 73.73 & 74.00&&\textcolor{gray}{\textit{61.53$_{-8.9}$}}&73.33$_{+2.9}$\\
\midrule
\midrule
\textbf{\textsc{AlignXplore}-7B}&\ding{51} & \textbf{61.97} &\textbf{65.33} & \textbf{68.53} & \textbf{67.59}&&\textbf{62.13}$_{-3.2}$&\textbf{71.27}$_{+8.6}$\\
\bottomrule
\end{tabular}
\end{adjustbox}
\label{evaluation_generalization}
\end{table}

\paragraph{Offline evaluation} Table~\ref{main_results} demonstrates the offline preference inference evaluation results. We draw six key findings: \textbf{(1) Necessity of preference inference:} Direct utilization of behavioral signals ($\mathcal{E}$) performs similarly to the ``Null'' setting and substantially worse than golden preference, validating the necessity of preference inference. \textbf{(2) Limitations of previous methods:} LMInductReason and VPL show poor performance, suggesting the inadequacy of prompting- and latent variable-based approaches. While PBA performs better through predefined preference modeling, its significant performance drop on \textsc{P-Soups} reveals limited generalization capability. \textbf{(3) Superiority of extended reasoning:} Models with extended reasoning consistently outperform their concise counterparts, as shown by Qwen3-32B$_\text{thinking}$ vs. Qwen3-32B$_\text{non-thinking}$ (65.03\% vs. 57.60\%) and DeepSeek-R1-671B vs. GPT-4 (70.47\% vs. 66.10\%). \textbf{(4) Strong performance of \textsc{AlignXplore} in base setting:} In the base setting, our model outperforms comparable-sized baselines on both in-domain and out-of-domain tasks, while achieving competitive performance with larger models like Qwen3-32B and GPT-4, even surpassing golden preference on \textsc{AlignX}$_\text{test}$. \textbf{(5) Dominant impact of RL:} While both stages contribute to performance, removing RL causes more significant degradation than removing cold-start training, indicating RL's critical role in optimizing preference alignment. \textbf{(6) Efficient and effective handling of growing signals:} When accessing additional historical information, DS-R1-Distill-Qwen-7B shows similar performance whether using more behavioral signals or leveraging previously inferred preferences. In contrast, our model significantly benefits from the streaming mechanism, outperforming even its 8-pair variant. This demonstrates that our streaming inference can more effectively utilize historical information through incremental preference refinement, while being more efficient than directly processing larger sets of behavioral signals. For simplicity, we adopt the base setting with $T=4$ for most subsequent experiments unless explicitly stated otherwise.


\paragraph{Online evaluation} Using GPT-4 as a judge for pairwise comparison of personalized response generation conditioned on the generated preference descriptions, Table~\ref{tab:win_rate} shows that \textsc{AlignXplore}-7B achieves competitive win rates against baselines on both in-domain and out-of-domain scenarios, further validating its effectiveness in preference inference.

\subsection{Generalization ability assessment}

We evaluate our model's generalization abilities from both input and output perspectives, as shown in Table~\ref{evaluation_generalization}.  \textbf{(1) Input-form generalization.} We evaluate models by replacing preference pairs with user-generated content (UGC) in the input signals, reflecting real-world scenarios where preferences must be inferred from diverse sources like reviews and social media posts. Our \textsc{AlignXplore}-7B exhibits strong generalization to different input formats, achieving 61.97\% accuracy that significantly outperforms baseline models.
\textbf{(2) Cross-model generalization.} We investigate the transferability of generated preference descriptions for personalizing different preference judging models, which is essential for broader adoption of preference-guided personalization systems. Our \textsc{AlignXplore}-7B demonstrates robust cross-model generalization, consistently outperforming baseline models of comparable size. We attribute this superior transferability to our extended reasoning mechanism, which encourages learning fundamental, model-agnostic preference patterns rather than surface-level correlations, resulting in more generalizable descriptions across different downstream models.

\begin{wrapfigure}[13]{r}{0.3\textwidth}
\vspace{-15pt}
  \centering
\includegraphics[width=0.25\textwidth]{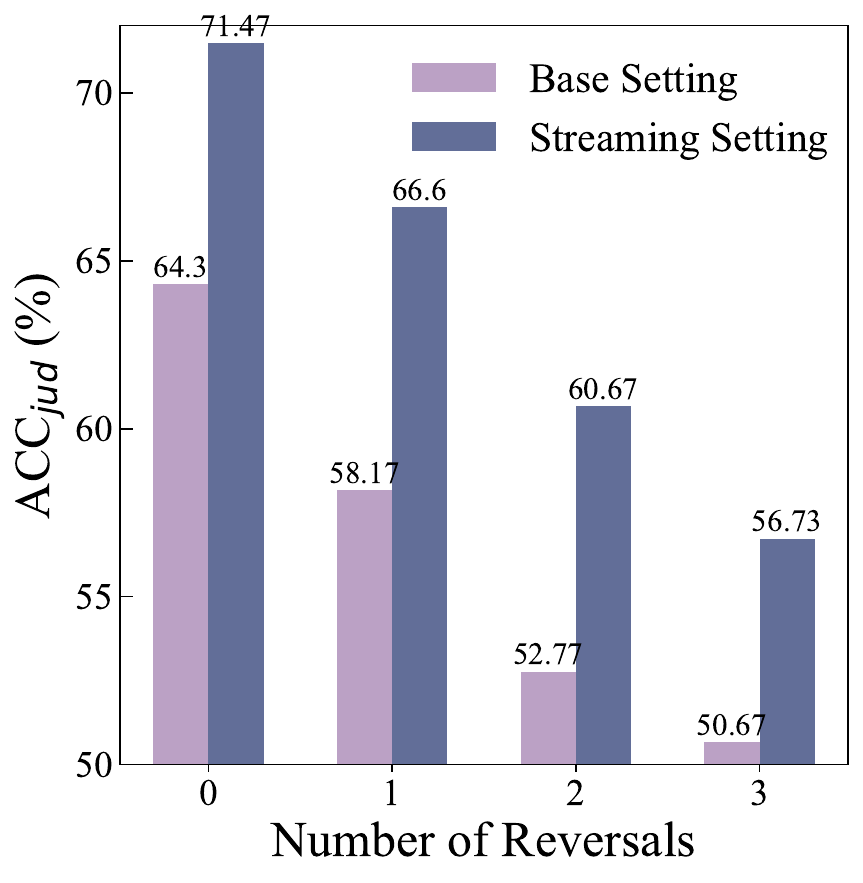}
\vspace{-10pt}
\caption{ACC$_\text{jud}$ of \textsc{AlignXplore}-7B on \textsc{AlignX}$_\text{test}$ as user preference reverses over time.}
  \label{fig: flip3}
\end{wrapfigure}

\subsection{Robustness assessment}

A key challenge for preference inference systems is maintaining consistent performance when user preferences differ significantly from training patterns. We first evaluate this robustness through preference reversal~\cite{li20251000000usersuserscaling}, where we reverse all preference directions in both behavioral signals and test pairs (e.g., changing ``$y_w\succ y_l$'' to ``$y_w\prec y_l$''). This tests whether the model truly learns to infer preferences rather than merely capturing fixed biases. As shown in Table~\ref{evaluation_generalization}, \textsc{AlignXplore}-7B demonstrates strong robustness with relatively small performance changes, notably outperforming both comparable-sized baselines and golden preferences. Even compared to DeepSeek-R1-671B, our model achieves competitive results, suggesting it learns to identify and adapt to preference patterns flexibly rather than relying on dataset biases.


We further investigate a more realistic and challenging scenario where user preferences evolve over time. Using 8 preference pairs from each user, we gradually reverse the preferences in earlier pairs while keeping the later pairs (and test pairs) consistent with the final preference. This simulates the common situation where historical behaviors may not fully align with current preferences. As shown in Figure~\ref{fig: flip3}, where the $x$-axis indicates the number of earlier signals with reversed preferences, our streaming inference mechanism consistently outperforms the base setting across different levels of preference shifts.\footnote{The robustness experiments on \textsc{P-Soups} are provided in Appendix~\ref{app: Robustness}.
} This demonstrates that explicitly modeling preference evolution through our streaming mechanism enables better adaptation to temporal preference changes, while methods that process all behavioral signals together may struggle to reconcile such inconsistencies.

\begin{wrapfigure}[15]{r}{0.5\textwidth}
\vspace{-15pt}
  \centering
  \includegraphics[width=0.5\textwidth]{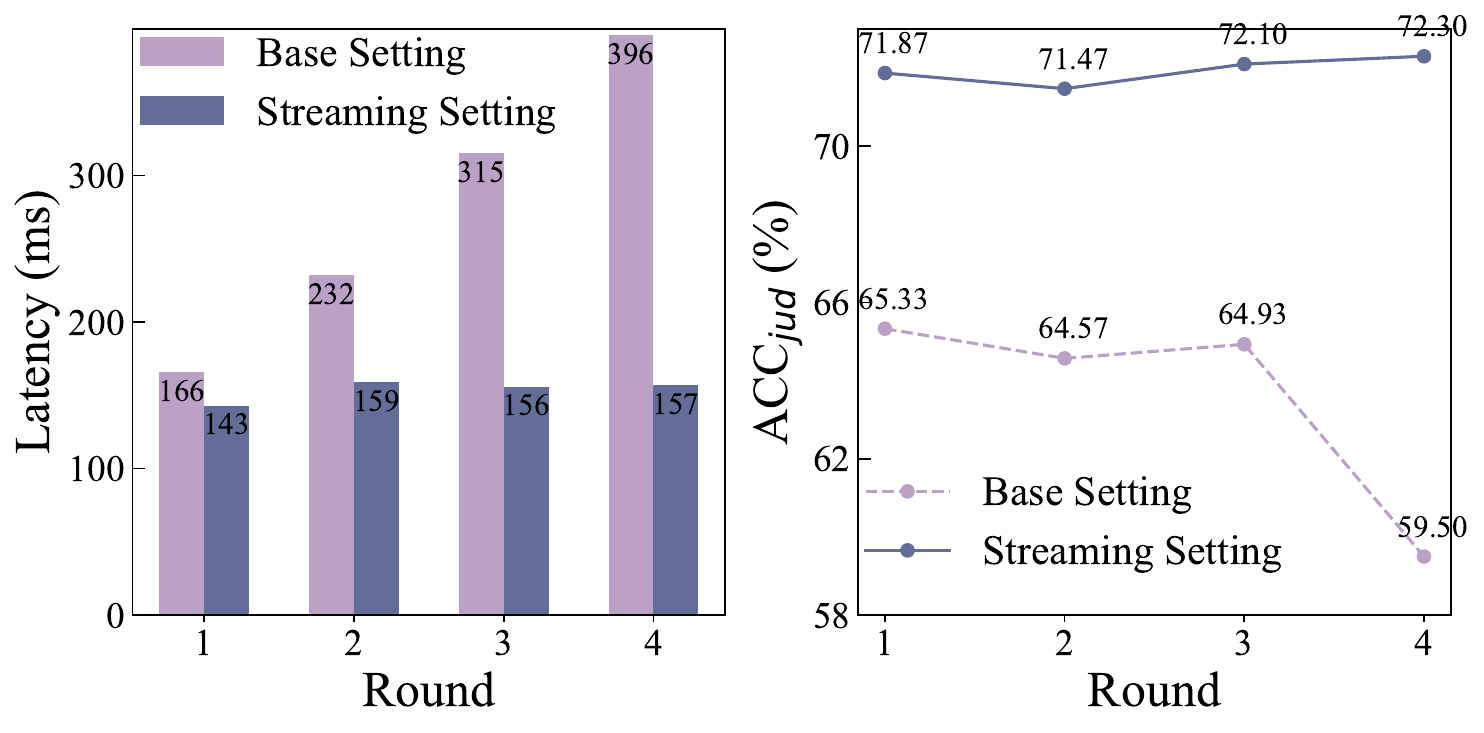}
\vspace{-15pt}
\caption{The average inference latency per example (\textbf{Left}) and ACC$_\text{jud}$ score (\textbf{Right}) of \textsc{AlignXplore}-7B on \textsc{AlignX}$_\text{test}$ as behavioral signals accumulate. Starting from 4 signals (Round=1), we add 4 new signals in each round. 
}
  \label{fig: latency}
\end{wrapfigure}

\subsection{Efficiency assessment}
\label{efficiency_assessment}


To evaluate computational efficiency as behavioral signals accumulate over time, we compare \textsc{AlignXplore}-7B's performance under base and streaming settings when processing increasing numbers of behavioral signals. As shown in Figure \ref{fig: latency}, we incrementally add 4 new behavioral signals in each round and measure the average inference time per example on \textsc{AlignX}$_\text{test}$. In the base setting, where all historical signals must be processed together, the inference time increases significantly with the number of signals. Moreover, when processing 16 signals (round 4), the model's performance (ACC$_\text{jud}$) drops sharply due to the challenge of handling long input contexts. In contrast, our streaming setting maintains stable inference time and performance by only processing the latest 4 signals along with the previously inferred preference description, demonstrating strong computational efficiency and performance for growing behavioral histories.

\begin{table}[t]
\centering
\small
\caption{Comparison of different reward functions for training preference inference models. $R_\text{jud}$ (used in all previous experiments) and $R_\text{gen}$ denote rewards from preference judging and response generation, respectively. \textbf{Takeaway:} $R_\text{jud}$ leads to better overall performance, improving both ACC$_\text{jud}$ and ACC$_\text{gen}$ despite optimizing only for judging accuracy. }
\begin{adjustbox}{max width=0.7\linewidth}
\begin{tabular}{lcccm{0.01em}cc}
\toprule
\multirow{2}{*}{\textbf{Method}} & \multirow{1}{*}{\textbf{Extended}} & \multicolumn{2}{c}{\textsc{\textbf{AlignX}}$_\text{test}$} && \multicolumn{2}{c}{\textsc{\textbf{P-Soups}}}\\
\cmidrule{3-4}
\cmidrule{6-7}
&\textbf{Reasoning}&\textbf{ACC$_\text{jud}$}&\textbf{ACC$_\text{gen}$} && \textbf{ACC$_\text{jud}$} & \textbf{ACC$_\text{gen}$}\\
\midrule
\midrule
${\mathcal{E}}$&N/A & 50.33 & 48.13 && 38.12 & 69.49\\
\rowcolor{gray!20}\textbf{Golden Preference}&N/A & \textcolor{gray}{\textit{64.63}} & 53.03 && 81.20 & 86.92\\
\midrule
\textbf{Qwen2.5-7B-Instruct}&\ding{55} & 56.33 & 48.53 && \underline{59.33} & \underline{72.22}\\
\textbf{DS-R1-Distill-Qwen-7B}&\ding{51} & 57.63 & 48.60 && 51.22 & 69.87 \\
\midrule
\rowcolor{gray!20}\textbf{DeepSeek-R1-671B}&\ding{51}& 70.47 & 50.65 && 70.44 & 80.42 \\
\midrule
\textbf{\textsc{AlignXplore}-7B ($R_\text{jud}$)}&\ding{51}& \textbf{65.33} & \underline{49.30} && \textbf{62.61} & \textbf{78.98}\\
\textbf{\textsc{AlignXplore}-7B ($R_\text{gen}$)}&\ding{51}& \underline{61.67} & \textbf{49.40} && 56.94 & 71.82\\
\bottomrule
\end{tabular}
\end{adjustbox}
\label{evaluation_reward}
\vspace{-10pt}
\end{table}

\begin{wrapfigure}[17]{r}{0.4\textwidth}
\vspace{-12pt}
  \centering
  \includegraphics[width=0.4\textwidth]{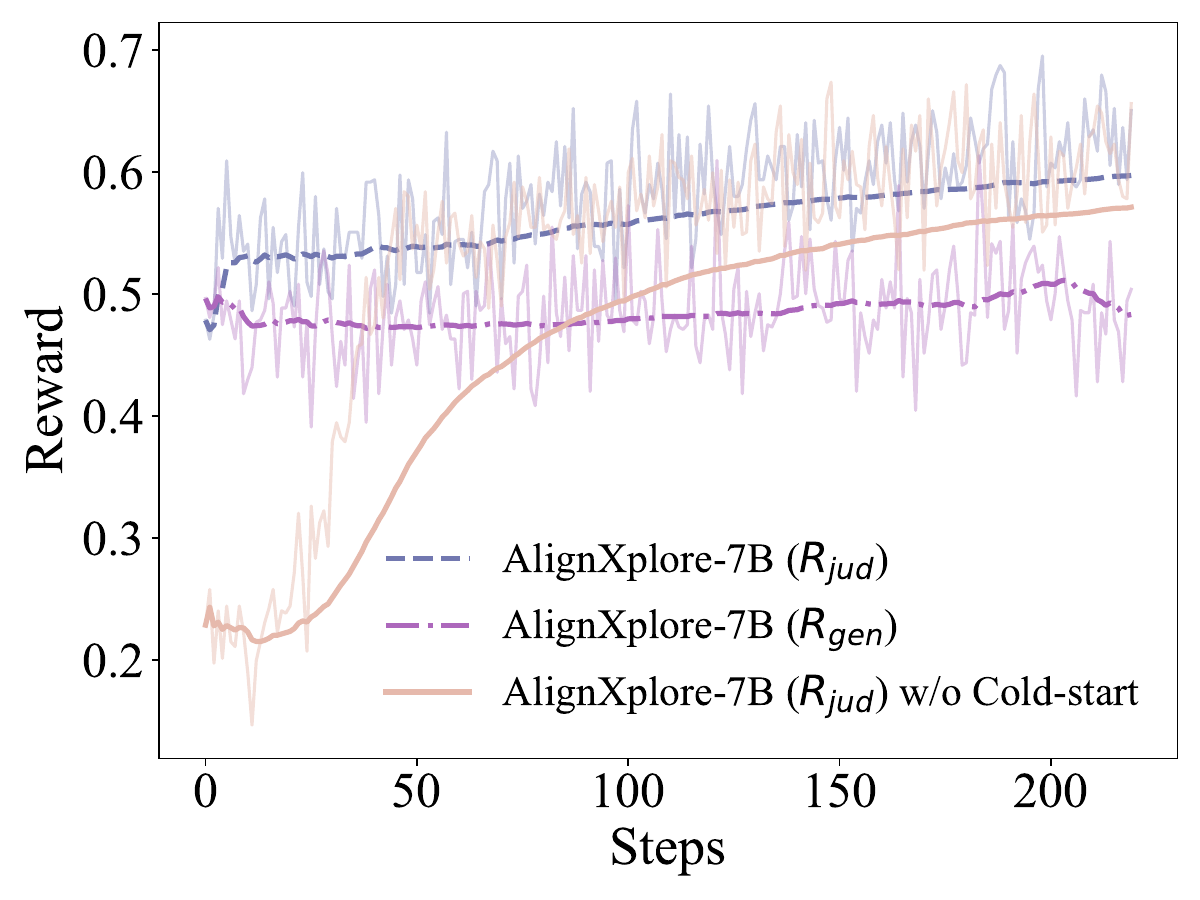}
\vspace{-17pt}
\caption{RL training curves of \textsc{AlignXplore}-7B with different reward functions. \textbf{Takeaway:} $R_\text{jud}$ provides more stable and effective training signals, showing consistent improvement over time, while $R_\text{gen}$ exhibits high variance and limited improvement.} 
  \label{fig: reward_curve}
\end{wrapfigure}

\subsection{Further analysis}
\label{sec: analysis}
Our further analysis focuses on two aspects: (1) comparing different reward functions (\textbf{Finding 1}), and (2) examining how two-stage training progressively enhances preference description quality (\textbf{Finding 2}). Additional analyses on RL training dynamics and detailed case studies are provided in Appendix~\ref{length_evolution} and~\ref{case_study}, respectively.



\paragraph{Finding 1: Optimizing for preference judging accuracy outperforms response generation rewards.} We investigate how different reward sources affect model performance by comparing $R_\text{jud}$ and $R_\text{gen}$, as shown in Table~\ref{evaluation_reward}. Results show that $R_\text{jud}$ achieves better performance across most metrics, even including response generation (ACC$_\text{gen}$), suggesting that accurate preference inference naturally facilitates better personalized generation.

We attribute this superiority to more informative training signals. As shown in Figure~\ref{fig: reward_curve}, $R_\text{jud}$ demonstrates steady improvement even without cold-start training, while $R_\text{gen}$ fluctuates around the random level (0.5). The ineffectiveness of $R_\text{gen}$ stems from two factors: (1) confounding factors in response probability estimation (e.g., language fluency, response length), and (2) inherently noisy reward computation from offline responses. While online examples might help, they require prohibitively expensive real-time user feedback. In contrast, $R_\text{jud}$ provides direct feedback about preference understanding, enabling stable training even from random initialization.

\begin{figure}[!t]
  \centering
\includegraphics[width=\linewidth]{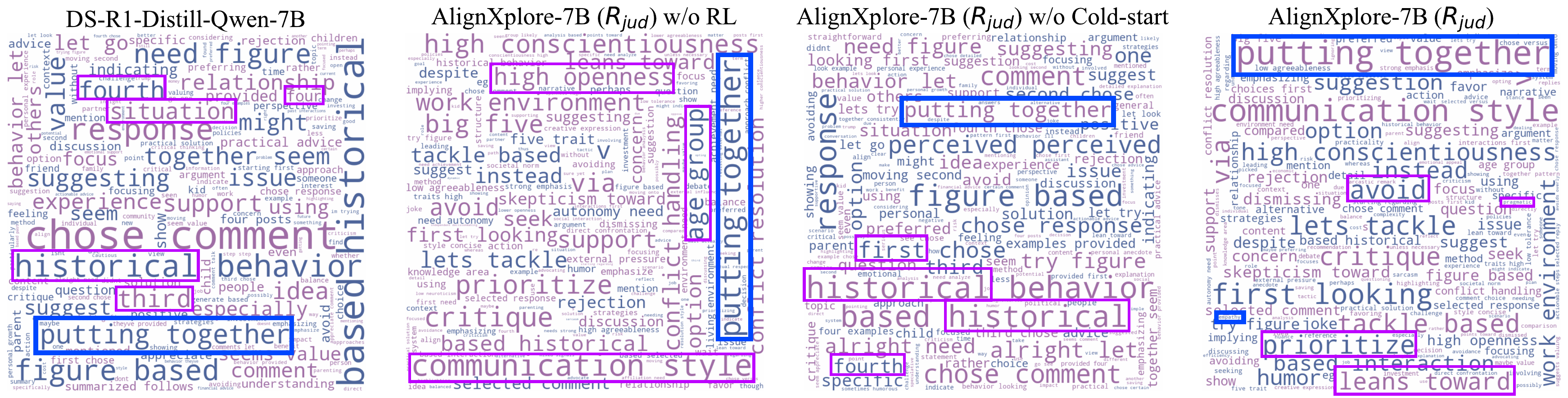}
  \caption{Word clouds of generated preference descriptions from model variants on \textsc{AlignX}$_\text{test}$. Terms in bounding boxes represent frequently occurring words characterizing each model's generation patterns. \textbf{Takeaway:} The evolution demonstrates how cold-start training helps identify preference dimensions, while RL learns to determine preference directions and aggregate signals across examples into actionable guidance, mirroring human inductive reasoning.} 
  \label{fig:wordcloud}
  \vspace{-15pt}
\end{figure}

\paragraph{Finding 2: Cold-start and RL training progressively enhance preference description quality.} Figure~\ref{fig:wordcloud} shows the word clouds of generated preference descriptions. The backbone model generates mainly general descriptions (e.g., ``historical,'' ``situation''). Cold-start training enables identification of specific preference dimensions (e.g., ``communication style,'' ``age group'') but shows limited synthesis capability. RL alone shows limited improvement, with manual inspection revealing a focus on generic dimensions (e.g., ``helpfulness''). Combining both stages leads to more actionable guidance with diverse preference dimensions and concrete actions (e.g., ``avoid,'' ``prioritize,'' ``leans toward''). This evolution mirrors human inductive reasoning~\citep{heit2000properties,FRANKEN2022101506}, evidenced by increasing usage of synthesis phrases like ``putting together'' during training. Without explicit supervision, our framework naturally encourages this iterative refinement process, progressively improving from general observations to specific, actionable preference hypotheses.

\section{Conclusion}

This work presents the first systematic investigation of extended inductive reasoning in LLMs through the lens of personalized preference inference. Our proposed model, \textsc{AlignXplore}, demonstrates that extended reasoning can effectively bridge the gap between implicit behavioral signals and explicit preferences. Through comprehensive experiments, we show that \textsc{AlignXplore} not only achieves substantial improvements in personalized alignment but also exhibits strong efficiency, generalization ability and robustness. The success of our two-stage training strategy provides valuable insights into developing LLMs' inductive reasoning capabilities, suggesting that combining synthetic demonstrations with reinforcement learning can effectively guide models to learn generalizable reasoning patterns rather than superficial correlations. Our findings also reveal several promising directions for future research, such as extending the success of our approach in preference inference to other inductive reasoning tasks, such as scientific hypothesis generation and pattern discovery in unstructured data.



\bibliographystyle{plain}
\bibliography{references}
\appendix




\section{Experiments}
\subsection{Implementation details}\label{implementation_details}
Our training and test sets are derived from \textsc{AlignX}, which proposes a 90-dimensional preference space (incorporating universal values, basic human needs, and prevalent interest tags). The dataset utilizes forum interactions and human-LLM interactions to construct 1.3 million examples, making it currently the largest and most comprehensive dataset for personalized alignment. However, preference signals in the original user interactions are relatively sparse, which previously hindered effective preference inference. To address this issue, we introduce a refined data construction approach. Specifically, we ensure that each target pair is associated with at least five preference dimensions, where all interaction history demonstrates consistent, non-neutral preference directions, while avoiding conflicting preferences across other dimensions. We constructed 10,000 data entries containing only ``pair-wise comparative feedback'' as interaction history, with 7,000 used for training and 3,000 for testing. When 7,000 instances are used for cold-start training, we select 3,980 instances for the first round and 5,278 instances for the second round based on $R(r, d) = 1$. Additionally, we constructed 3,000 entries containing only ``user-generated content'' as interaction history for generalization validation.

The training is conducted on 8 NVIDIA A100 GPUs using Adam optimizer~\citep{kingma2014adam}, with DeepSpeed ZeRO-3~\citep{rajbhandari2020zero} and Flash-attention-2~\citep{dao2023flashattention} for optimization. We employ the following hyperparameter configuration: learning rate of 1e-6, 50 warmup steps, 4 training epochs, and maximum prompt/generation lengths of 8,192/2,048 tokens. During RL, we set the mini-batch size to 128 for each step.

\subsection{Baseline details}\label{baseline_details}

We compare our approach with various baseline methods and models: 
\begin{itemize}[leftmargin=10pt]
    \item \textbf{Directly given preference descriptions:} (1) \textit{Null}: no preference description is provided; (2) $\mathcal{E}$: using behavioral signals directly as preference descriptions without inference; and (3) \textit{Golden Preference}: ground-truth preference descriptions provided by the benchmark. Note that golden preference descriptions, while semantically accurate, may not necessarily lead to optimal downstream personalization performance due to potential gaps in model compatibility.
    \item \textbf{Previous specialized methods for inductive reasoning and personalization:} (1) \textit{LMInductReason}~\cite{qiu2024phenomenal} employs iterative hypothesis refinement to enhance LLMs' inductive reasoning capabilities; (2) \textit{VPL}~\cite{poddar2024personalizing} introduces latent variables to model individual preferences; and (3) \textit{PBA}~\cite{li20251000000usersuserscaling} maps behavioral examples to structured preference scores along predefined dimensions, then converts them to natural language descriptions.
    \item \textbf{Preference descriptions generated by state-of-the-art LLMs:} The LLMs range from small-sized models including \textit{Qwen2.5-7B-Instruct}~\cite{qwen2.5} and \textit{DS-R1-Distill-Qwen-7B}~\cite{deepseekai2025deepseekr1incentivizingreasoningcapability}, to large-sized models including \textit{QwQ-32B}~\cite{qwq32b}, \textit{Qwen3-32B}~\cite{qwen2025qwen3}, \textit{GPT-4}~\cite{achiam2023gpt}, and \textit{DeepSeek-R1-671B}~\cite{deepseekai2025deepseekr1incentivizingreasoningcapability}. These models cover both concise reasoning and extended reasoning patterns.
\end{itemize}
Furthermore, to verify the effectiveness of our approach, we also compare with \textsc{AlignXplore}-7B w/o RL and w/o Cold-start under the base setting, which only uses cold-start training and RL for preference inference, respectively.

For VPL~\cite{poddar2024personalizing}, we train one epoch on Qwen2.5-7B-Instruct using $\mathcal{D}_\text{rl}$. Note that this method employs its own specialized downstream model for preference-guided judgment. For other baselines, we generate roles or preferences using the corresponding models and input them into Qwen2.5-7B-Instruct for evaluation. LMInductReason~\cite{qiu2024phenomenal} follows the original paper's implementation, where content generation is replaced by Qwen2.5-7B-Instruct. After iteratively generating rules, the final rule is provided to Qwen2.5-7B-Instruct to generate preference selections. PBA~\cite{li20251000000usersuserscaling} uses the method from the original paper to extract consistent preferences from the interaction history of each benchmark.

\subsection{Length evolution}\label{length_evolution}

\begin{wrapfigure}[12]{r}{0.45\textwidth}
\vspace{-15pt}
  \centering
\includegraphics[width=0.45\textwidth]{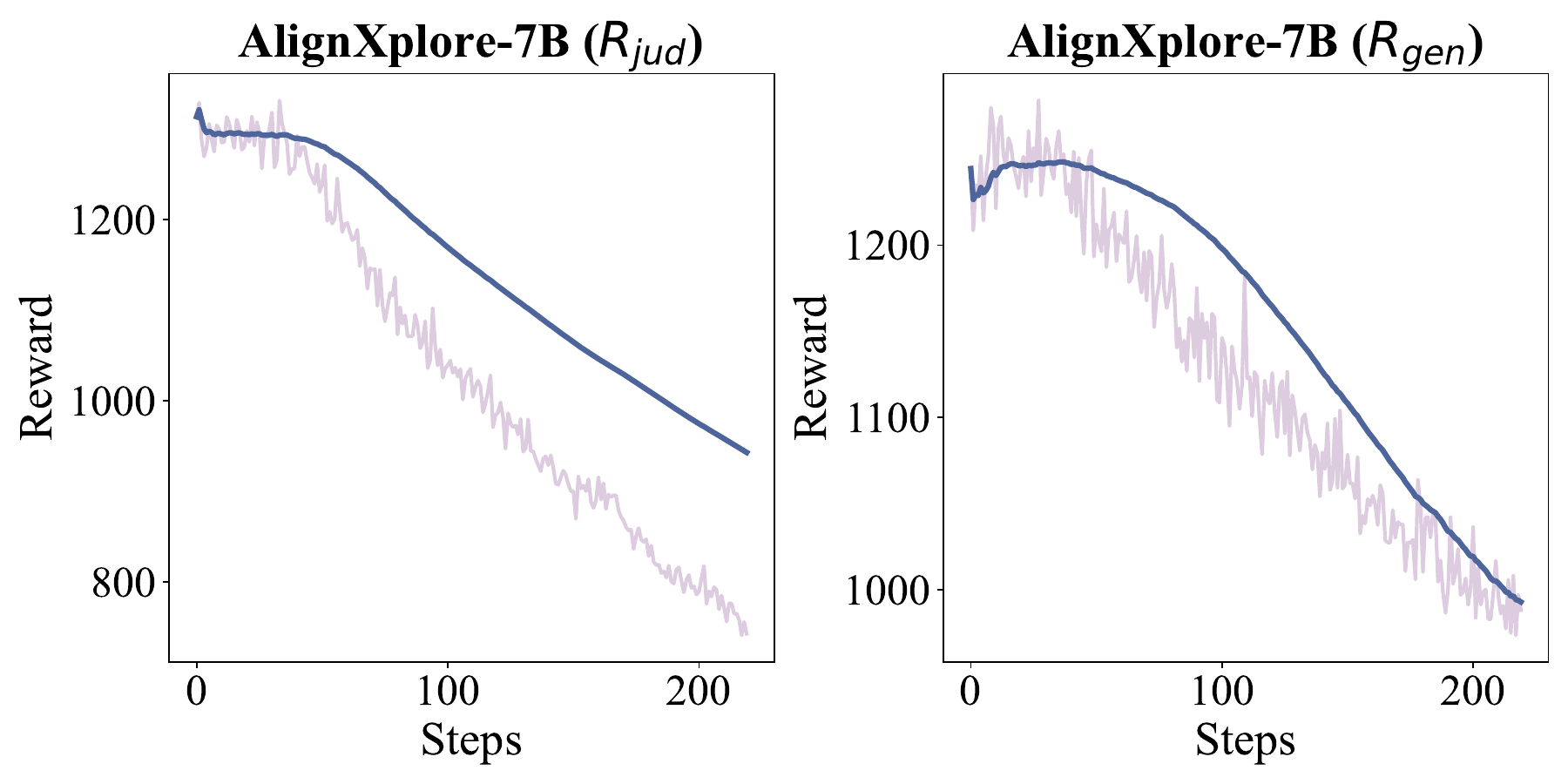}
\vspace{-20pt}
\caption{Curves of generation length for \textsc{AlignXplore}-7B with different reward functions during RL training.}
  \label{fig: length}
\end{wrapfigure}

We present the changes in generation length during the reinforcement learning process for \textsc{AlignXplore}-7B ($R_\text{jud}$) and \textsc{AlignXplore}-7B ($R_\text{gen}$) in Figure \ref{fig: length}. As training progresses, the average generation length of the model continuously decreases. Our analysis suggests that, due to cold-start training, although the model is guided to analyze the appropriate preference dimensions, it tends to repetitively reproduce content from the behavioral signals, with low confidence in the analysis and many redundant and fluctuating dimensional interpretations. After reinforcement learning, the model's analysis direction becomes clearer. For preference interpretation of behavioral signals, the model now only mentions key terms that reflect preferences, enabling it to quickly analyze and summarize user preferences. This aligns with the analysis presented in \S\ref{sec: analysis}.

\subsection{Robustness assessment}\label{app: Robustness}

\begin{table}[t]
\centering
\small
\caption{ACC$_\text{jud}$ of \textsc{AlignXplore}-7B when reversing the first preference pair of the user and keeping the later pairs (and test pairs) consistent with the final preference. \textbf{Extended Reasoning}: whether the model generates preference descriptions with extended reasoning. $T$ refers to the number of examples in $\mathcal{E}$ in both training and inference. $\hat{d}$ indicates whether historical preferences are empty or not.}
\begin{adjustbox}{max width=\linewidth}
\begin{tabular}{lcccccccc}
\toprule
\multirow{2}{*}{\textbf{Method}} & \multirow{1}{*}{\textbf{Extended}} &\multirow{2}{*}{\textbf{Setting}} &\multirow{2}{*}{$T$} & \multirow{2}{*}{$\hat{d}$} & \multirow{2}{*}{\textsc{\textbf{AlignX}}$_\text{test}$} & \multicolumn{3}{c}{\textsc{\textbf{P-Soups}}}\\
\cmidrule{7-9}
&\textbf{Reasoning}&&&&&\textbf{Informativeness} & \textbf{Style} & \textbf{Expertise} \\
\midrule
\midrule
\textbf{\textsc{AlignXplore}-7B}&\ding{51}&Base&8&\ding{55} & 58.17&51.66&67.67&61.17\\
\textbf{\textsc{AlignXplore}-7B}&\ding{51}&Streaming&4&\ding{51} & 66.60 & 58.97 & 69.67 & 65.67\\
\bottomrule
\end{tabular}
\end{adjustbox}
\label{apptab: Robustness}
\end{table}

Figure \ref{apptab: Robustness} shows the performance of \textsc{AlignXplore}-7B on different datasets under the base setting and streaming setting when 8 preference pairs are given for each user and the first preference pair is reversed. Since the streaming inference mechanism allows the model to refine preference descriptions during streaming inference of user preferences, it demonstrates robustness and generalization when facing inconsistent or time-varying user behavior preferences. As a result, it consistently outperforms the base setting on both in-domain and out-of-domain datasets.

\subsection{Case study}\label{case_study}
DS-R1-Distill-Qwen-7B tends to be more general and one-sided when analyzing preferences from behavioral signals, which may lead to the omission of important points during the analysis. After cold-start training, \textsc{AlignXplore}-7B w/o RL provides more comprehensive and systematic analysis of the preference dimensions, but expressions indicating uncertainty, such as ``?'' and ``Not clear yet,'' frequently appear, along with extensive repetitions of content from the behavioral signals, such as ``User describes facing harassment by a host due to his identity.'' After reinforcement learning, these are replaced by more confident statements and clearer analyses, indicating that RL significantly aids in making inductive reasoning more precise and focused. In the streaming setting, behaviors that contrast with and adjust according to historical preferences may emerge, such as ``looking at the past preferences,'' ``fits,'' and ``consistently.''

For non-extended-reasoning models (e.g., Qwen2.5-7B-Instruct), the preference descriptions are provided directly. However, due to the lack of reasoning processes, some unreasonable preference descriptions emerge. In fact, during the analysis, the model focuses more on the user's responses or the user's tendencies toward different responses, rather than focusing on the content of the questions. However, many of the analyses provided by Qwen2.5-7B-Instruct are based on the content of the questions, such as ``Interest in Personal Development and Self-Improvement.''

\begin{tcolorbox}[colframe=grey!20!white, colback=grey!3!white, coltitle=black, fonttitle=\bfseries, title=Prompt for Case Study, breakable]

A conversation between User and Assistant. The User asks a question, and the Assistant solves it. The Assistant first thinks about the reasoning process in the mind and then provides the User with the answer. The reasoning process is enclosed within \texttt{<think>} \texttt{</think>} and answer is enclosed within \texttt{<answer>} \texttt{</answer>} tags, respectively, i.e., \texttt{<think> reasoning process here </think>} \texttt{<answer> answer here </answer>}.

\textbf{User:} You must put your answer inside \texttt{<answer>} \texttt{</answer>} tags, i.e., \texttt{<answer> answer here </answer>}.

\textbf{This is the problem:} \\
Generate the user's preference based on their historical behavior.

\textbf{This person has chosen or rejected comments on some posts:}

\begin{enumerate}
    \item \textbf{Post:} \\
    Sorry for format on mobile etc. \\
    My girlfriend[22] and I[22] decided to go away somewhat last minute. It\textquotesingle{}s our first trip together. We\textquotesingle{}re away in France, not far from Lille. We decided to get an apartment on Airbnb, it was inexpensive and so beautiful. It was perfect. \\
    Except, and this is the very concise version, we show up and the host looks surprised to see us. I speak fluent french and my girlfriend doesn\textquotesingle{}t, and he doesn\textquotesingle{}t speak any English so I held the conversation even though she handled all the booking and liaising. He kept on asking if my girlfriend was the one in the picture as he was expecting the male and female in the picture. Her picture was an old one of her and her friend. I tried explaining this to him and he acts as if I\textquotesingle{}m not understanding french, my own first language, properly. \\
    Turns out that my girlfriend had arranged for a bouquet for me which was waiting for me in the main room. He couldn\textquotesingle{}t wrap his head around it. He was acting somewhat civilised until I saw the flowers and hugged my girlfriend. Then he pieced it together and started acting hostile. His tone changed and he started laying down impromptu ground rules like no liquids near the sofa and no shoes on the rug and don\textquotesingle{}t trash the place, return it how you got it. All while his tone got more and more hostile. \\
    It\textquotesingle{}s his apartment and he probably has a key and I just feel so unsafe. I\textquotesingle{}ll get over it once I process it I\textquotesingle{}m sure but I\textquotesingle{}m literally just starting medication for depression and anxiety and being at home is so comforting for me, which is why I was looking forward to getting in after travelling for what felt like eternity. It just fucking sucks that I can\textquotesingle{}t go away with my significant other without feeling unsafe and vulnerable. And I feel so bad cus my girlfriend is trying so hard to make this the best trip and she feels bad that our host is a homophobe even tho that\textquotesingle{}s literally nothing to do with her. \\
    Idk if this made sense I\textquotesingle{}m tired and sleep deprived but can\textquotesingle{}t bring myself to sleep. Please send some kind gay words so I don\textquotesingle{}t feel like the only gay in the world. Little pick me up please.
    
    \textbf{Chosen:} \\
    I live 20mins from Lille! (In Belgium near the Belgian/French border) \\
    I\textquotesingle{}ve heard before that people near the border/north of France seem to be more conservative. \\
    I\textquotesingle{}m sorry this happened to you.
    
    \textbf{Rejected:} \\
    Un bon gros fdp en somme.
    
    \item \textbf{Post:} \\
    So, I\textquotesingle{}m a 21yo bisexual (maybe pansexual? I\textquotesingle{}m completely open to it but haven\textquotesingle{}t really met or talked to someone to know), living in the Chicago area. I realized and have been out as bisexual since I was 17. I\textquotesingle{}ve slept with women, and I hate to admit it because of the 5 I\textquotesingle{}ve been with, only 2 were legitimately into girls, the rest was just drunken young adult barsexual experimenting. This is already sounding great right? \textit{*sarcasm*} Well, I\textquotesingle{}ve dated my fair share of men and quite honestly I\textquotesingle{}m bored. I have yet to find a girl that there is a mutual interest. 2 or 3 girls have liked me, but I honestly just wasn\textquotesingle{}t into them in a romantic way, and that was all they wanted, no friendship. I really want to date a girl already! I have no clue how to flirt with girls, I can never tell when I\textquotesingle{}m being flirted with, or how to tell them I\textquotesingle{}ve been out as bisexual for almost 5 years, have never dated a girl but screwed a bunch. Doesn\textquotesingle{}t sound too good on my end. I kind of have a crush on a bisexual friend of mine and would like to try for her, but I\textquotesingle{}ve tried and failed at flirting or catching her attention in more than a friendly way. I have no clue what to do unless I\textquotesingle{}m in bed with them. I have no idea how to flirt with girls. Send help.
    
    \textbf{Chosen:} \\
    I wish I could offer some good advice, but I\textquotesingle{}m not so good at flirting myself. \\
    I wish you the bestest of luck.
    
    \textbf{Rejected:} \\
    I flirt like a pro and I can be smoother than butter... But I have no idea how to articulate the HOW aspect of it. I really want to help, but I\textquotesingle{}m grasping for words here. \\
    \textit{*code-sloth ponders a moment, as a sloth does when stumped*}
    
    Witty banter and humor are my main skills, which take a bit to learn for each gal\textquotesingle{}s tastes. At the very least, be genuine and sincere. Whether you\textquotesingle{}re a smooth talker or as mentally coordinated as a deer on ice, sincerity goes a long way to endearment. Thoughtful, sincere, genuine. Start there, then you can develop your own style of flirting. \\
    Oh! And be aware of when to shut up or back off. That great one-liner she lined you up for? Don\textquotesingle{}t always blurt it out. She\textquotesingle{}s not reacting well to certain things (usually dirty lines)? Cool your jets and mellow out. It\textquotesingle{}s a balancing act, and you\textquotesingle{}ll pick up on it over time. \\
    Sometimes you won\textquotesingle{}t even need to flirt actively. Talking about a mutual subject (I love video games, for example) can cause the "oh, I dig this chick" feeling that flirting does. I\textquotesingle{}m not above the whole "Can this controller fit between your boobs?" line of discussion, but you don\textquotesingle{}t have to be on your witty toes ALL the time. Would you date a good yet incompatible flirt or someone you shared a common interest with? Probably the latter. \\
    Regarding pickup lines: No. Don\textquotesingle{}t use them in serious context. Jokingly yes, but don\textquotesingle{}t play that card on the table first. \\
    Woah, that got a bit verbose. Sorry!
    
    \item \textbf{Post:} \\
    As a 20 year old, it made me sad to see so many of you calling yourself old! Not that that\textquotesingle{}s a bad thing. I don\textquotesingle{}t think teenage/20s years are the peak of your life. I was having this conversation with my ex girlfriend (yeah...I know) the other day and she said this is a really shitty confusing time and IA. and besides I have so many health issues, I\textquotesingle{}m looking forward to having surgery and stabilising and having more of a grip on my life/mental stability in my 30\textquotesingle{}s and 40\textquotesingle{}s and all the years after that. \\
    Anyway this thread is kind of OT but to all the 30+ ladies on here, you\textquotesingle{}re not old and even when you are you can still be a badass. I know 70 year olds that are kicking ass and when you think about it they were born in the 1940\textquotesingle{}s, and healthcare sucked then. \\
    Saw a thread here asking for members over 30...
    
    \textbf{Chosen:} \\
    Thanks, thats nice of you to say :)
    
    \textbf{Rejected:} \\
    It\textquotesingle{}s cool. I\textquotesingle{}m immortal anyway. Which reminds me, I have this sword fight I need to get to... \\
    \textit{*Queen Plays*}
    
    \item \textbf{Post:} \\
    My boss, after ghosting me for two weeks, fired me over text this morning. She was fine when I came out to her, but after discovering that I\textquotesingle{}d actually take medication to change my body, she\textquotesingle{}s hated me. I worked with kids so she was always afraid I\textquotesingle{}d corrupt their minds. As far as I know, parents didn\textquotesingle{}t care or know. They assumed me to be female and their kids assumed me to be male but otherwise it was all fine. \\
    I guess I\textquotesingle{}ll stay with my other job and wait for them to start paying me (still a volunteer until the fall when a job opens up). At least I\textquotesingle{}m out there and everyone cool with it (behavioral health center). \\
    Should I just stay closeted in the future? I guess I will until I pass and then I\textquotesingle{}d be stealth? Fired for "inability to keep up with changes"
    
    \textbf{Chosen:} \\
    When are we going to get laws protecting LGBT people from getting fired? How come we didn\textquotesingle{}t get it yet? I\textquotesingle{}m so sorry this happened to you.
    
    \textbf{Rejected:} \\
    If your employer has not been documenting these "failures," then they are not in a good position legally. The EEOC (equal employment opportunity commission) is a federal agency that investigates discrimination due to gender identity. You can and should contest your firing; check them out online today. You have two weeks from firing to contest. Please do. \\
    It\textquotesingle{}s possible that nothing will come of it, but very often, they can find other workers who will substantiate your manager\textquotesingle{}s motives. You will have done the right thing by putting it into their hands. \\
    Obviously, you have to assess your own emotional needs before taking this step. \\
    Quick action is also important for unemployment, which is the right of any worker released for no fault of their own. You will have to contest the firing with the unemployment office, too. Your manager is cheating you financially. \\
    Keep your head up. If you choose, this is new work, more important than any you have done before. Your task is to try to get justice for yourself and other trans people. Just to try. That is more than enough. \\
    It\textquotesingle{}s great of you to volunteer at the behavioral health center! I hope your career there will be rewarding.
\end{enumerate}

\textbf{Assistant:} \texttt{<think>}

\end{tcolorbox}

\begin{tcolorbox}[colframe=grey!20!white, colback=grey!3!white, coltitle=black, fonttitle=\bfseries, title=Preference Generated by DS-R1-Distill-Qwen-7B, breakable]

Okay, so I need to figure out the user's preference based on their historical behavior. They've provided four examples where they either chose or rejected comments on posts. Let me analyze each one step by step.

\textbf{First Post:} The user shared a personal anecdote about traveling with a partner and facing a negative host due to language and cultural barriers. The user felt unsafe and lost depression and anxiety. They received a chosen comment that was empathetic and a rejected one that was dismissive. From this, it seems like the user appreciates genuine support and understanding, especially in sensitive situations.

\textbf{Second Post:} The user identified as a bis (or pan) and discussed dating experiences. They felt unsuccessful and frustrated. The chosen comment was a friendly acknowledgment, while the rejected one was more detailed but still empathetic. The user values honesty and being heard, especially in personal and vulnerable contexts.

\textbf{Third Post:} The user shared about facing termination due to their gender identity, feeling misunderstood and unhelped. The chosen comment was a compliment, and the rejected comment was more informative but still empathetic. The user seems to thrive on positive reinforcement and feels supported when others acknowledge their strengths.

\textbf{Fourth Post:} Another situation where the user faced professional discrimination. They were fired for being closeted and chose a comment that expressed solidarity and disappointment, while the rejected comment provided practical steps for legal action. The user values solidarity and stands with others in similar situations but also seeks practical advice.

Looking at these, the user seems to prefer comments that are empathetic, supportive, and acknowledge their struggles. They appreciate genuine concern and positive reinforcement. They also seem to seek practical advice when applicable, especially in situations where they feel discriminated or unsafe. However, they don't seem to appreciate overly dismissive or critical comments, even if they're well-meaning.

So, the user's preference is for comments that show understanding, support, and positivity. They value empathy and practical help when needed, but they don't like being dismissed or told to "get over it." They likely seek validation, reassurance, and encouragement in their challenging situations.

\texttt{</think>} \\

The user's preferences are evident in their analysis of past interactions, where they favor comments that demonstrate empathy, support, and positivity. They appreciate genuine concern and acknowledge the user's struggles, seeking both practical advice and affirmation in their challenging situations. They dislike overly dismissive or critical comments, even if well-intentioned, and value solidarity and encouragement. Thus, the ideal comment for the user would be one that shows understanding, offers support, and provides positive reinforcement or practical advice when applicable.

\end{tcolorbox}

\begin{tcolorbox}[colframe=grey!20!white, colback=grey!3!white, coltitle=black, fonttitle=\bfseries, title=Preference Generated by \textsc{AlignXplore}-7B w/o RL, breakable]

Okay, let\textquotesingle{}s start breaking down the information provided here. The task is to generate the user\textquotesingle{}s preferences based on their historical interactions—specifically the comments they\textquotesingle{}ve chosen versus rejected across different social media posts. There are several factors mentioned related to psychological needs and communication styles mentioned earlier, such as openness, agreeableness, autonomy, approach to conflict resolution, etc., along with specific communication nuances.

First, looking at each example provided:

\textbf{Starting with Post \#1:} \\
User describes facing harassment by a host due to his identity. His main issue here seems emotional distress stemming from societal norms affecting his identity, particularly gender-related discrimination. He feels unsafe, uses medication for mental health issues, seeks support, and wants validation. 

He chose a response acknowledging location near a border area (which might indicate regional context?), noting potential conservatism elsewhere, expressing sympathy, and thanking him. Rejected a vague remark (\textit{un bon gros fdp en somme}), suggesting the user prefers empathetic responses rather than dismissive ones.

So possible indicators include High Openness (to experience/emotional intelligence?), Low Agreeableness since he\textquotesingle{}s upset, High Need for Autonomy—he wants control over his situation (e.g., staying home, dealing with anxiety independently?), Conflict Handling – confrontational or avoidance? Not clear yet.

\textbf{Moving onto Post \#2:} \\
User identifies as bisexual (maybe pan), struggling with attraction despite previous experiences. Struggles finding authentic romantic partners beyond casual encounters. Wants advice but feels frustrated. Chooses a reply offering sympathy and encouragement ("bestest luck"), rejects a lengthy, verbose guide filled with witty tips. That suggests preferring brevity over detail-heavy solutions—maybe valuing conciseness vs thorough explanations.

Possible clues: Low openness to complex strategies, higher need for autonomy in handling situations themselves versus relying on external methods. Also, rejects overly structured approaches preferring direct empathy.

\textbf{Third Interaction (Post\#3):} \\
User celebrates aging positively, mentions achieving maturity, health challenges, and future goals. Chose a simple thank-you response instead of playful/humor-oriented replies referencing swords/fighting. Indicates seriousness regarding aging positively, dislikes sarcasm or flippant replies.

Shows emphasis on sincerity/respectful communication avoiding trivialization. Emphasizes authenticity and substance over levity.

\textbf{Fourth Scenario (Boss situation):} \\
Emotionally impacted by losing job over gender identity concerns. Selected compassionate question about legislation protection plus sympathy (“how come?”), whereas the rejected comment gives legal/procedural guidance including steps to contest via agencies and unemployment offices. The user chooses emotional expression over procedural advice. Suggests prioritizes emotional support and communal solidarity over practical steps. Prioritizes feeling validated and supported rather than facing bureaucracy head-on immediately.

So here autonomy comes in – user prefers staying informed but seeks immediate comfort/support rather than taking proactive steps immediately. Maybe prefers empathetic validation over practical action plans.

\textbf{Now compiling these observations:}

\textbf{Looking at Personality Traits:}

\begin{itemize}
    \item \textbf{Big-Five Openness:} Lower probably. Because they reject verbose/detailed answers in favor of shorter/more straightforward suggestions. Unless the rejection is specific to inappropriate content... Not sure though. Alternatively, openness relates to creativity/exploration—maybe limited here.
    
    Wait, but in Post \#1, the rejection of “un bon gros fdp en somme” (which roughly translates to \textquoteleft{}f*** off\textquoteright{}) shows aversion to crude language, preferring respectful tone. Maybe mid-openness? Hmm tricky.
    
    \item \textbf{Agreeableness:} Likely higher than average given tendency to choose comforting/empathic remarks over confrontational or sarcastic ones. However, in fourth instance, the chosen comment asks \textquoteleft{}how come\textquoteright{} implying empathy but also acknowledges systemic issues (laws?), indicating balance between supportive and critical perspectives? Or perhaps moderate agreeableness.
    
    \item \textbf{Need For Autonomy:} Seems strong—they prefer independent solutions, resisting dependence on others\textquotesingle{} input (as seen in second post rejecting elaborate tips). Though in fourth scenario, they sought help from authorities (law/companies?) but still chose emotional affirmation over procedural steps. Maybe balancing autonomy with seeking expert resources selectively.
    
    \item \textbf{Conflict Handling Approach:} From first and fourth cases, the user reacts strongly emotionally (anxiety, anger), indicating avoidance of direct confrontation unless necessary. Preferred responses offer reassurance rather than addressing root causes directly. Thus prefer indirect/empathic resolutions.
    
    \item \textbf{Communication Style:} From examples above, the user tends toward concise messages avoiding excessive details. Rejected options included lengthy guides filled with anecdotes/examples. Hence leans towards Concise/Prefer direct communication over elaboration.
    
    \item \textbf{Work Environment Preferences:} Uncertain yet. Third post mentions volunteering positively. Fourth talks about employment issues. Possibly values environments allowing flexibility and autonomy, avoiding rigid structures that led to firing. Prefers places where boundaries respected ethically (maybe corporate policies aligned with LGBTQ+ rights?).
    
    \item \textbf{Workplace Needs:} Likely values transparent workplaces adhering to anti-discrimination laws, thus advocating for legal protections. Also wants safe spaces free from hostile environments (as per first story).
    
    \item \textbf{Achievement Needs:} In third scenario celebrates maturity and long-term objectives (ageing positively), showing appreciation for progress and stability. Achievement motivation exists but framed positively (not competitive, but attaining inner peace/stability).
    
    \item \textbf{Autonomy Needs:} Strong. Avoid relying on others\textquotesingle{} input (rejected tips in \#2), prefers making own decisions guided by empathy (e.g., trusting their instincts in dating despite uncertainties). Want control over decisions, reduce dependency on external validation/suggestions.
    
    \item \textbf{Law Awareness:} Fourth scenario acknowledges potential lack thereof affecting workplace fairness. Suggests awareness/careful consideration regarding legislation. May value informed decisions backed by legal insights but balance with emotional responses.
\end{itemize}

\textbf{Putting together all threads...}

\textbf{Finalizing categories:}

\textbf{Big Five:}
\begin{itemize}
    \item \textbf{Low/Openness?}: Since rejects complex advice, leans toward simpler expressions. However, openness relates to imagination/intellect—not sure. Maybe neutral/mid-range?
    
    \item \textbf{Moderate-High Agreeableness}: Chooses supportive tones, avoids hostility. Empathetic nature evident throughout choices.
\end{itemize}

\textbf{Needs:}
\begin{itemize}
    \item \textbf{Strong Autonomy}: Prefers independent judgment, resists external influences (rejection of structured guides in dating, seeking validation instead of proactive measures).
    \item \textbf{Need for Achievement}: Mixed signals—but focuses on overcoming anxieties via proactive actions (getting back mentally).
    \item \textbf{Workplace Safety}: Prioritizes environments respectful of identities, advocating policies preventing discrimination.
    \item \textbf{Conflict Management}: Prefer mediating through dialogue/acknowledging issues indirectly rather than combative methods.
\end{itemize}

\textbf{Communication Style:} Concise, avoids unnecessary elaboration. Focus on clarity and sincerity.

\textbf{Law:} Shows awareness of systemic factors impacting employment/freedom, willing to challenge injustices but seeking respectful dialogues.

Thus synthesizing all points into coherent preferences...

\texttt{<think>} \\

\texttt{<answer>}

Based on the analyzed behaviors, the user exhibits the following preferences derived from their interactions:

\begin{enumerate}
    \item \textbf{Personality Traits (Big Five):}
    \begin{itemize}
        \item \textbf{Openness:} Moderately low/high. Prefers pragmatic solutions over elaborate suggestions (rejects verbose advice but values empathy).
        \item \textbf{Agreeableness:} High. Chooses supportive, compassionate language and prioritizes interpersonal harmony.
    \end{itemize}
    
    \item \textbf{Core Psychological Needs:}
    \begin{itemize}
        \item \textbf{Need for Autonomy:} Very pronounced. Resists reliance on external advice/influences (rejects complex dating tips, favors personal intuition). Seeks self-determined pathways amidst stress/anxiety.
        \item \textbf{Workplace Safety:} Strong advocacy for inclusive workplaces free from bias/discrimination.
        \item \textbf{Emotional Security Over Pragmatism:} Values emotional validation and communal solidarity (choosing heartfelt condolences) over pragmatic steps (legislation guidance).
    \end{itemize}
    
    \item \textbf{Social Interaction Styles:}
    \begin{itemize}
        \item \textbf{Directness + Empathy:} Communicates concisely yet sincerely, balancing honesty with sensitivity (chooses brief affirmation vs verbose explanations/rejections).
        \item \textbf{Avoidant Conflict Resolution:} Tends to sidestep contentious topics (e.g., avoids discussing workplace discrimination head-on except when validating feelings).
    \end{itemize}
    
    \item \textbf{Legal Awareness \& Personal Achievement:}
    \begin{itemize}
        \item A conscious awareness of legal frameworks influencing life decisions (aware of EEOC protections but leans on communal solidarity).
        \item Personal fulfillment tied to overcoming vulnerabilities (mental health improvement linked to traveling away for escape).
    \end{itemize}
    
    \item \textbf{Work Environment Needs:}
    \begin{itemize}
        \item Thrives in transparent, equitable workplaces aligned with individual values (supports LGBTQ+ rights legislation advocacy).
    \end{itemize}
\end{enumerate}

\texttt{</answer>}

\end{tcolorbox}

\begin{tcolorbox}[colframe=grey!20!white, colback=grey!3!white, coltitle=black, fonttitle=\bfseries, title=Preference Generated by \textsc{AlignXplore}-7B under Base Setting, breakable]

Okay, let's start breaking down these interactions to figure out the user's preferences using the given criteria. First, looking at each post response:

\textbf{Starting with Post 1:} The user chose a comment acknowledging the situation politically ("people near the border") and showing empathy. They rejected a shorter phrase ("un bon gros fdp"). This suggests the user values contextual analysis and empathy rather than brevity. So maybe they prefer deeper insights over concise replies.

\textbf{Moving to Post 2:} Here, they picked a supportive reply emphasizing luck and sincerity without practical advice. Rejected a verbose explanation filled with tips. This points toward preferring brevity and warmth over extensive suggestions. Also, rejecting humor ("witty banter") implies less interest in overtly playful solutions unless sincere.

\textbf{In Post 3:} Chose gratitude instead of sarcastic remarks about immortality. Shows appreciation for genuine thanks versus edgy content. Emphasizes gratitude and positivity.

\textbf{Lastly, Post 4:} Opted for emotional support asking for societal protections vs professional guidance steps. Indicates valuing emotional validation first before actionable advice. Rejects detailed procedural guidance, focusing on feelings.

Putting together, the user likely prioritizes warmth, empathy, conciseness, and avoids overly verbose/narrative-heavy responses. Their communication leans towards direct yet compassionate messages, seeking connection through shared experiences over complex explanations. They might value authenticity and emotional support above technical details or sarcasm.

\texttt{</think>} \\

\texttt{<answer>} \\
\textbf{User Preferences Analysis Based on Historical Behavior:}

\textbf{Key Traits Identified Across Selected Responses:}

1. \textbf{Emotional Support Over Detailed Advice:}
   - Prefers empathetic affirmations (e.g., "Thanks, that\textquotesingle{}s nice of you" vs lengthy guides on legal processes).

2. \textbf{Contextual Humor/Social Awareness:}
   - Chooses politically relevant observations (border regions) over blunt humor ("un bon gros fdp").

3. \textbf{Direct Empathy Without Overexplanation:}
   - Selects compassionate statements focusing on connection ("be sincere"), rejects verbose explanations (flirting tips, "let me help").

4. \textbf{Privacy Focus:}
   - Avoids overly personal/critical content, opting for platonic solidarity (volunteering at health centers, "stealth" strategy).

\textbf{Personality Traits Alignment:}
- Low openness to abstract concepts (preferring straightforward empathy).
- High agreeableness (prioritizing harmonious interactions).
- Strong desire for \textbf{authentic communication}, disliking pretentious language or performative politeness. \\
\texttt{</answer>}

\end{tcolorbox}

\begin{tcolorbox}[colframe=grey!20!white, colback=grey!3!white, coltitle=black, fonttitle=\bfseries, title=Preference Generated by Qwen2.5-7B-Instruct, breakable]

Based on the user's historical behavior on the platform, the following preferences and tendencies can be inferred:

\begin{enumerate}
    \item \textbf{Support for LGBTQ+ Community:}
    \begin{itemize}
        \item The user frequently engages with posts and comments related to the LGBTQ+ community, especially regarding trans issues, relationships, and personal experiences.
        \item They often choose to offer comfort, support, and validation to others going through similar struggles, showing empathy and a supportive nature.
    \end{itemize}
    
    \item \textbf{Interest in Personal Development and Self-Improvement:}
    \begin{itemize}
        \item The user seeks advice and guidance on personal issues such as dealing with ex-partners, coming out, and navigating family and social relationships.
        \item They are open to receiving and giving advice, showing a willingness to engage in meaningful conversations that can help others.
    \end{itemize}
    
    \item \textbf{Emotional Support:}
    \begin{itemize}
        \item The user often chooses comments that provide emotional support, validation, and encouragement. They appreciate kind words and genuine responses.
        \item They tend to avoid confrontational or dismissive comments, preferring to offer understanding and empathy.
    \end{itemize}
    
    \item \textbf{Sensitivity to Pronouns and Gender Issues:}
    \begin{itemize}
        \item The user demonstrates sensitivity to pronoun usage and gender issues, often choosing comments that reflect understanding and support for trans individuals.
        \item They are likely to be aware of and respectful of different gender identities and pronouns.
    \end{itemize}
    
    \item \textbf{Struggle with Intimacy and Relationships:}
    \begin{itemize}
        \item The user frequently mentions difficulties with intimacy, relationships, and sexual experiences, indicating a need for support and guidance in these areas.
        \item They seem to be seeking validation and advice on how to navigate relationships, both romantic and platonic.
    \end{itemize}
    
    \item \textbf{Interest in Sexual Health and Consent:}
    \begin{itemize}
        \item The user shows an interest in sexual health and consent, discussing sensitive topics such as sexual experiences and boundaries.
        \item They value explicit and honest communication in relationships, showing a preference for clear and respectful interactions.
    \end{itemize}
\end{enumerate}

Based on these preferences and tendencies, it can be inferred that the user values emotional support, empathy, and genuine advice in their interactions. They are likely to appreciate content that is respectful, understanding, and supportive of their struggles, especially in relation to personal relationships, gender issues, and sexual health.

\end{tcolorbox}

\begin{tcolorbox}[colframe=grey!20!white, colback=grey!3!white, coltitle=black, fonttitle=\bfseries, title=Preference Generated by \textsc{AlignXplore}-7B under the Streaming Setting, breakable]

Okay, let me start breaking this down. The task is to generate the user's preferences based on their selected versus rejected comments across these four posts. First, I need to analyze each post individually, noting why the chosen response stands out compared to the ones they dismissed.

Starting with Post 1: Here, the author talks about dealing with a homophobic host and feeling unsafe. Their chosen comment is brief—mentioning living nearby and pointing towards regional attitudes without much emotional support. Rejected was ``Un bon gros fdp en somme.'' Hmm, maybe the rejection indicates preferring shorter responses instead of longer French phrases. Also, the chosen comment shows understanding via location-related insight rather than direct empathy.

Moving to Post 2: A bisexual guy struggling with dating. Chosen reply offers sympathy (``bestest'') briefly, whereas the longer, detailed rejection includes advice and humor. Maybe shorter responses resonate better, avoiding verbose explanations. Or perhaps the user prefers conciseness despite the topic complexity.

Third Post discusses aging positively. Selected reaction is simple gratitude, rejected uses dark humor (``immortal''). Choosing the simple thank-you suggests preferring straightforward replies over sarcastic takes.

Fourth Post involves workplace discrimination. User picked a brief complaint asking for legal protections, rejecting a practical legal guide filled with steps. Again, shorter, emotional statements preferred over informative/detailed guidance.

Looking at patterns:
\begin{itemize}
  \item \textbf{Conciseness}: All chosen replies are shorter, direct messages. Reject longer, helpful options.
  \item \textbf{Empathy Focus}: Chosen responses often show shared experiences (e.g., location note implying regional bias) or brief solidarity, not offering solutions.
  \item \textbf{Avoidance of Sarcasm/Judgment}: Some rejected answers use dark humor or harsh tones.
  \item Short sentences preferred over lengthy ones except when necessary.
\end{itemize}

Personality-wise, maybe high agreeableness (preferring compassionate responses), low openness to complex advice. Prefers brevity over detail, dislikes long form suggestions. Emotionally, seeks empathetic validation quickly rather than getting bogged down in practicalities.

So summarizing:
Preferences lean towards succinct, empathetic communication avoiding unnecessary elaboration. Prefer concise replies highlighting common issues rather than deep analysis. Dislikes lengthy guidance or sarcasm, gravitates toward brief affirmations.

</think>\\

<answer>

User favors \textbf{concise, emotionally attuned responses} over elaborate or directive ones. Key preferences include:
\begin{enumerate}
  \item \textbf{Brevity}: Chooses shorter, punchier remarks (e.g., ``I live 20 mins...'', ``Thanks!'') over lengthier explanations.
  \item \textbf{Empathy Over Detail}: Prioritizes relatable context (regional bias, shared struggles) rather than offering structured solutions.
  \item \textbf{Avoidance of Sarcasm/Judgment}: Rejected responses contained irony or negativity (dark humor, legal jargon-heavy advice).
  \item \textbf{Need for Quick Validation}: Prefers brief affirmations (``Thanks'', ``Sorry'') over prolonged discussions.
\end{enumerate}

These preferences suggest a tendency for \textbf{concise communication}, valuing immediacy and vulnerability, while shunning complexity or prolonged guidance.

</answer>

\end{tcolorbox}

\begin{tcolorbox}[colframe=grey!20!white, colback=grey!3!white, coltitle=black, fonttitle=\bfseries, title=Prompt for Updating the 4 Behavior Signals under the Streaming Setting, breakable]

A conversation between User and Assistant. The User asks a question, and the Assistant solves it. The Assistant first thinks about the reasoning process in the mind and then provides the User with the answer. The reasoning process is enclosed within <think> </think> and answer is enclosed within <answer> </answer> tags, respectively, i.e., <think> reasoning process here </think> <answer> answer here </answer>. User: You must put your answer inside <answer> </answer> tags, i.e., <answer> answer here </answer>.

This is the problem:
Summarize user preferences based on their behavior. If past preferences are provided, adjust the preferences by combining past preferences with those reflected in current behavior, removing conflicting parts, and integrating new insights. If no past preferences are provided, derive the final preferences solely from user behavior.

\textbf{Past Preferences:}

User favors \textbf{concise, emotionally attuned responses} over elaborate or directive ones. Key preferences include:
\begin{enumerate}
  \item \textbf{Brevity}: Chooses shorter, punchier remarks (e.g., ``I live 20 mins...'', ``Thanks!'') over lengthier explanations.
  \item \textbf{Empathy Over Detail}: Prioritizes relatable context (regional bias, shared struggles) rather than offering structured solutions.
  \item \textbf{Avoidance of Sarcasm/Judgment}: Rejected responses contained irony or negativity (dark humor, legal jargon-heavy advice).
  \item \textbf{Need for Quick Validation}: Prefers brief affirmations (``Thanks'', ``Sorry'') over prolonged discussions.
\end{enumerate}

These preferences suggest a tendency for \textbf{concise communication}, valuing immediacy and vulnerability, while shunning complexity or prolonged guidance.

\textbf{This person has chosen or rejected comments on some posts:}

\begin{enumerate}
    \item \textbf{Post:} This is just a vent and, of course, it's not directed to my lovely spouse, since she's supportive and great. Still, it bothers me so much that people have such a hard time respecting my pronouns, yet as soon as she came out, everybody started using her name and pronouns correctly (she's MtF). We're both at the same stage in our transitions (pre-everything) and when she came out, I could see the immediate change in her friends. She only found out later in life that she's trans, while I've been struggling with it my whole life. It's as if just because I was AFAB, my transition somehow has to be taken less seriously. I don't know, maybe I'm exaggerating. I'm just pretty bummed, mates. My pronouns hardly get respected, yet everyone respects my wife's.

    \textbf{Chosen:} Dealing with the exact same, actually. Literally the same. Down to the last letter. Sorry brother.
    
    \textbf{Rejected:} I always seek solace in the fact that those who misgender me are going to look absolutely dumb one day.

    \item \textbf{Post:} she said she is gonna support me if i think im trans (in her words) for much longer, but said im not allowed a haircut because i will hate myself more apparently. how the fuck do i even react to this? she makes these random suggestions based off of her own knowledge rather than fact and wont let me correct her. she said i never shower signs of when i was younger, but i did, she just never noticed it. i dont know what to do. i came out to my mum and she said im going through a phase.

    \textbf{Chosen:} mine said to my father "you know how SHE is, SHE was just venting" when I came out to them. you know what’s good for you, nobody else does. she just needs time to understand it. hold on bro
    
    \textbf{Rejected:} Go get your hair cut anyway. Any friends can bring you?

    \item \textbf{Post:} So I am bi.  Have always been more attracted to women than men.  Sometimes I wonder am I gay?  Who knows.  I\'ll figure it out. Some background about me: I haven\'t been able to meet a woman to date long term.  I have mostly only hooked up with women one night stand style, or friends with benefits. I can easily meet men, but I always meet women simultaneously. Any guy I am with has to be 100\% ok with that.  The women I have really really crushed on or fell in love with either moved away or just dropped off the face of the earth and stopped answering my calls.  (the reason for this back story is because I think I would have figured out an answer to my question if I had had a long term gf.) I met a woman the a few weeks ago and we hit it off.  I didn\'t want to have sex with her right away, even though I could tell she was hinting toward it.  I wanted to have a chance to explain to her what I enjoy and have a mutually pleasurable experience. I told her I had to end the night and that I would like to have dinner with her in a few days.  We had dinner, drinks, more drinks, and by this time we were both pretty tipsy and I felt comfortable enough to flirt and tell her that if we ever got together that I need a lot of warm up.  I thought I was pretty clear in explaining that I have a VERY sensitive clit, but we got back to my place and I kissed her, and and then all of a sudden, she is violently mashing my vulva with her face and aggressively fingering me.  I just kind of ended it because I said I had too much to drink.  I wasn\'t enjoying it at all and I felt a little violated after I specifically told her the parameters of my body. So. Here\'s the thing!  This happens to me. A lot.  I don\'t really understand why.  Even with a disclaimer of "I need to warm up with lots of touching and kissing and teasing and I need a feather light touch to get me going." Do I have the most sensitive clit ever had on a woman\'s body?  Is it more common to just mash the vulva around and suck on a clit with full force? Do most women enjoy just one kiss and then straight to deep aggressive fingering?

    \textbf{Chosen:} Not alone. I\'m not fond of direct touching on my clit, I\'d rather someone rub it through the folds around it. I\'ve had girlfriends who have consistently ignored this, went straight for the clit every time, and then got pissed when I just squirm around uncomfortable while they\'re giving me, in their opinion, the best sex ever.
    
    \textbf{Rejected:} No, you sound totally normal. Your problem isn\'t sensitivity, it\'s that you\'re not being assertive enough. Instead of saying "oh, I had too much to drink let\'s stop", you need to tell your partner "hey, that\'s way too fast/too hard. You need to slow down." Don\'t even take your panties off if you\'re not ready for that, just tell her "not yet, kiss me here", etc. If she gets mad or refuses to listen, end it there (but don\'t lie about the reason).

    \item \textbf{Post:} A little rant. My dad said that to me recently. I came out to him like 2 years ago. He said he was ok with me being gay after a week of not talking to me after I came out to him. What I\'m wondering is how he\'d feel when I\'d get a girlfriend (I never had dated yet. I just turned 19 in September). Would having a girlfriend be too shove-y for him? I don\'t know. I wish he could just tell me straight up that he isn\'t really that accepting. He is a conservative so I guess I shouldn\'t be surprised. A lot of people since lgbt started becoming more accepting have said stuff like that a lot. What are you\'re experiences when it comes to people who say these types of phrases and how do you feel about it? [Rant] "I\'m ok with gay people ... Just don\'t shove it in my face"

    \textbf{Chosen:} When I came out, everyone but one sister "accepted me."  Now, every single family member has blocked me on Facebook, except my two gay uncles.  One sister-in-law has completely cut me out to the point of not allowing me near their children, only exception being family events like my mom\'s recent birthday.  Saddest part was when my little 4-year-old nephew asked me, "why haven\'t I seen you in so long???"  I had to change the subject while fighting back tears. Thank god for my gay uncles.  I\'m going to make a point to visit them ASAP.  *hugs* for everyone going through similar rough patches with toxic people.
    
    \textbf{Rejected:} I hate people like that. Its just a bullshit double standard. Heterosexuals *constantly* do the same things that, when done by gay couples, are "shoving it in the face," and nobody even blinks. Can you imagine the shitstorm that would go down if even half of LGB people started telling straight people to stop shoving it in our faces whenever they did things like hold hands in public?
    
\end{enumerate}

Assistant: <think>

\end{tcolorbox}

\begin{tcolorbox}[colframe=grey!20!white, colback=grey!3!white, coltitle=black, fonttitle=\bfseries, title=Preference Generated by \textsc{AlignXplore}-7B after Updating the 4 Behavior Signals under the Streaming Setting, breakable]

Okay, so I need to summarize the user's preferences using the given framework. Let's see... First, looking at the past preferences:

They prefer concise, empathetic responses avoiding sarcasm or judgment. They value quick validation without long explanations. So, their communication style leans towards brief and heartfelt messages rather than lengthy ones.

Looking at the examples provided:

In Post 1, the user chose a response that empathizes (``dealing with the same...''), which fits brevity and shared struggle. Rejected the sarcastic comment which uses dark humor.

For Post 2, the selected reply offers comfort and practical advice (``nobody needs to rush understanding'') whereas the rejected option suggests a humorous question, perhaps seen as dismissive.

On Post 3, the preferred reaction validates the experience (mentioning discomfort with direct contact) without giving specific advice. The rejected suggestion pushes active correction, which might feel abrupt.

Lastly, in Post 4, the accepted message shares personal relatable pain succinctly, whereas the rejected one introduces a broader critique which could extend beyond the immediate situation.

So overall, the user prefers responses that acknowledge shared feelings, validate their situation, and avoid pushing solutions or jokes. They avoid anything sarcastic or overly directive. Their needs align with concise empathy, avoiding negativity, and seeking immediate connection.

<think>\\

<answer>

The user consistently favors \textbf{concise, emotionally resonant statements} that validate shared experiences and foster vulnerability, shunning humor, sarcasm, or prescriptive advice. Key traits observed include:
\begin{enumerate}
  \item \textbf{Brevity}: Selects brief, relatable anecdotes (e.g., ``Same situation''). Rejects verbose/detailed suggestions.
  \item \textbf{Empathy-driven validation}: Chooses responses acknowledging pain (e.g., ``Saddest part was when my little 4-year-old nephew...'') instead of analytical critiques.
  \item \textbf{Avoidance of judgment/negative framing}: Dismisses sarcastic remarks (``dumb''), ironic questions, or framing problems as moral failures.
  \item \textbf{Value for brevity}: Prefers unadorned affirmations (``Thanks!'') over elaborated guidance.
\end{enumerate}

These choices reflect a preference for \textbf{empathetic, unstructured solidarity} over structured problem-solving, emphasizing immediacy and emotional authenticity in communication.

</answer>

\end{tcolorbox}

\section{Data format and prompt}
\label{appendix:prompt}

\begin{tcolorbox}[colframe=grey!20!white, colback=grey!3!white, coltitle=black, fonttitle=\bfseries, title=Evaluation Prompt, breakable]

Determine which response the user prefers based on the user\textquotesingle{}s preferences. Please output your selection below in a json format by filling in the placeholders in []:
\begin{verbatim}
{"selection": "[Response A / Response B]"}
\end{verbatim} \\

\textbf{Prompt} \\
It is just recently that I realized I have no idea what the job market for chemists looks like, although I am a chemistry student. Could someone shed some light on that? Also, which of those bring in the most money? \\
Chemistry jobs? \\

\textbf{Preference} \\
\textbf{This person's persona is:} \\
This person likes science, knowledge. An aspiring chemistry professional who values optimism and factual information, preferring to gather insights from collective experiences rather than individual anecdotes. This person is proactive in seeking career guidance and appreciates straightforward, encouraging responses. \\

\textbf{Response A} \\
I just had a job interview with a private analytical company a week ago. They were small and privately owned. You could have considered it an entry level position. They weren't expecting anyone to be FULLY trained in their analytical methods, which were not very common. I found out after I DIDN'T get the job that I had to compete with two Ph.D.s and a ton of other people. I'm fresh out of undergrad. Twas my first job interview and I am glad I had the experience of getting that for the next few job interview experiences I have possible coming up, but wow, I was getting screened for that job while two other doctorate fellows were also totally into it. \\

\textbf{Response B} \\
I think the people in this thread are a bit pessimistic. Someone did a salary thread a few weeks ago and it didn't look bad at all. One thing a prof mentioned to me is that companies pretty much will not hire someone without lab experience as other people have mentioned. \\

\begin{verbatim}
{"selection": "Response B"}
\end{verbatim}

\textbf{Prompt} \\
\{prompt\} \\

\textbf{Preference} \\
\{persona\} \\

\textbf{Response A} \\
\{responseA\} \\

\textbf{Response B} \\
\{responseB\}

\end{tcolorbox}

\begin{tcolorbox}[colframe=grey!20!white, colback=grey!3!white, coltitle=black, fonttitle=\bfseries, title=Prompt for Generating Reasoning Chains and Preference Descriptions under Base Setting, breakable]

Generate the user's preference based on their historical behavior.

The following aspects can be referred to when analyzing user preferences.

\{key preference dimensions\} \\

\textbf{This person has chosen or rejected comments on some posts:} \\
\{implicit preference signals\}

\end{tcolorbox}

\begin{tcolorbox}[colframe=grey!20!white, colback=grey!3!white, coltitle=black, fonttitle=\bfseries, title=Prompt for Generating Reasoning Chains and Preference Descriptions under Streaming Setting, breakable]

Summarize user preferences based on their behavior. If past preferences are provided, adjust the preferences by combining past preferences with those reflected in current behavior, removing conflicting parts, and integrating new insights. If no past preferences are provided, derive the final preferences solely from user behavior.

The following aspects can be referred to when analyzing user preferences.

\{key preference dimensions\} \\

\textbf{Past Preferences:}

\{past preferences\} \\

\textbf{This person has chosen or rejected comments on some posts:} \\
\{implicit preference signals\}

\end{tcolorbox}

\section{Limitations}
\label{Limitations}
Due to the lack of a real LLM-user interaction test platform, we were unable to validate the model's reasoning performance in a real-world environment. Once such a testbed becomes available, we will evaluate our model's performance on it. This paper primarily focuses on the scenario of preference inference, ensuring that the historical preferences in the test set are consistent with the test pairs. Future work could extend to scenarios where user preferences change dynamically over time, requiring the model to adjust preferences based on the user's recent behaviors during inference.

\section{Impact statement}
\label{Impact}
This work enhances the preference inference capability of models, enabling them to better serve human users by understanding and responding to their individual preferences. However, it may involve potential risks related to user privacy and bias. By inferring personalized preferences, there is a possibility of inadvertently amplifying existing biases in the data or misinterpreting user intent. To mitigate these risks, we ensure that our approach incorporates robust fairness and transparency measures. We also prioritize user consent and implement mechanisms to ensure that user data is anonymized and securely handled. Furthermore, we encourage ongoing monitoring of the model\textquotesingle{}s performance in real-world scenarios to identify and address any unintended consequences, thus ensuring that the model's deployment remains ethical and aligned with user interests.

\end{document}